\documentclass{article}

\PassOptionsToPackage{numbers, compress}{natbib}

 \usepackage[main, final]{neurips_2026}

\usepackage[utf8]{inputenc}
\usepackage[T1]{fontenc}
\usepackage{hyperref}
\usepackage{url}
\usepackage{booktabs}
\usepackage{amsfonts}
\usepackage{amsmath}
\usepackage{amssymb}
\usepackage{mathtools}
\usepackage{bm}
\usepackage{nicefrac}
\usepackage{microtype}
\usepackage{xcolor}
\usepackage{wrapfig}
\usepackage{enumitem}

\newcommand{\C}{\mathbb{C}}
\newcommand{\R}{\mathbb{R}}
\newcommand{\Imag}{\mathrm{Im}}
\newcommand{\Real}{\mathrm{Re}}
\newcommand{\by}{\mathbf{y}}
\newcommand{\bz}{\mathbf{z}}
\newcommand{\bx}{\mathbf{x}}
\newcommand{\br}{\mathbf{r}}

\newcommand{\bo}{\mathbf{o}}
\newcommand{\bone}{\mathbf{1}}
\newcommand{\bzero}{\mathbf{0}}
\newcommand{\bgamma}{\bm{\gamma}}
\newcommand{\bomega}{\bm{\omega}}
\newcommand{\balpha}{\bm{\alpha}}
\newcommand{\blambda}{\bm{\lambda}}
\newcommand{\bdelta}{\bm{\delta}}
\newcommand{\bW}{\mathbf{W}}
\newcommand{\bA}{\mathbf{A}}
\newcommand{\bB}{\mathbf{B}}
\newcommand{\bC}{\mathbf{C}}
\newcommand{\bD}{\mathbf{D}}

\newcommand{\bI}{\mathbf{I}}
\newcommand{\bLambda}{\mathbf{\Lambda}}
\newcommand{\chit}{\chi_\tau}

\title{Prospective Coding Improves Learning in Deep Continuous-Time Recurrent Networks}

\author{Shivang Rawat$^{1, *}$ \hspace{14pt} Mirko Morello$^{1}$ \hspace{14pt} Flaviano Morone$^{1, 2}$ \hspace{14pt} David J. Heeger$^{1, 2}$ \vspace{10pt} \\
$^1$ Telepath \\
$^2$ New York University\\
\\
$^*$\texttt{shivang@telepath.com}
}

\begin{document}

\maketitle

\begin{abstract}
Temporal integration gives continuous-time recurrent networks memory, but in deep stacks it also delays bottom-up signals and attenuates top-down errors. We compensate for this delay by making each layer's bottom-up input prospective, and we develop \textit{Recursive Quadrature Filters} (RQFs), biologically motivated temporal band-pass filters that are a special case of diagonal state-space models (SSMs). Starting from an energy model, we derive the RQF dynamics and show that each RQF is a band-pass filter whose learnable parameters control its tuning frequency and bandwidth. We then make each layer's bottom-up input prospective using a parameter-free two-tap update that leaves the recurrent transition and parallel scan unchanged. We extend this correction to general diagonal SSMs and show that it mitigates depth-dependent gradient attenuation when temporal gradients are truncated, i.e., spatial-only backpropagation. We evaluate the performance of RQFs, S5, and ORGaNICs (a nonlinear gated RNN with built-in attentional modulation and recurrent normalization) trained using full backpropagation through time (BPTT) and spatial-only backpropagation. Under full BPTT, the prospective variants of these models match or outperform their non-prospective controls in every configuration.
A non-residual width-32 six-layer RQF achieves $96.09\pm0.23\%$ accuracy on raw-audio Speech Commands with $31.9$k parameters; a width-64 six-layer RQF achieves $83.36\pm1.71\%$ on the $16{,}384$-step Path-X task. We also introduce selective RQFs (sRQFs), which make each filter's input gain and, optionally, its bandwidth and tuning frequency input-dependent while retaining the parallel scan. With the input gain alone, the same six-layer model achieves $86.31\pm0.51\%$ on Path-X; with all three input-dependent, a four-layer prospective sRQF solves Selective Copying ($99.83\pm0.07\%$). These results identify RQFs as a parameter-efficient recurrent substrate and prospective-input coding as an input-side correction for deep continuous-time recurrent networks.
\end{abstract}

\section{Introduction}
\label{sec:intro}

Biological neural circuits face a fundamental trade-off between memory and reactivity. A biological neuron is well modeled as a leaky integrator whose membrane time constant $\tau$ confers temporal context but also delays and attenuates fast input changes~\citep{gerstner2014neuronal}. When stacked into an $L$-layer feedforward circuit, this filter cascades, so the effective relaxation time grows roughly as $L\tau$~\citep{haider2021latent}. The same trade-off is built into recurrent theories of cortical function~\citep{heeger2017theory}, in which slow recurrent dynamics let the network settle into a representation that combines feedforward drive with prior expectation. This lag also creates a learning problem: when a top-down teaching signal is applied at the output layer, the activity it should instruct in early layers is already roughly $L\tau$ out of date. Signals that vary on timescales comparable to $\tau$ are therefore poorly served by the standard leaky cascade.

This paper contributes a recurrent substrate and an input-side correction. The substrate is the \textit{Recursive Quadrature Filter} (RQF), a linear dynamical system inspired by neuroscience that is a special case of a diagonal state-space model (SSM), so it trains on sequences of tens of thousands of time steps with standard SSM algorithms. The input-side correction drives each layer with a prospective rather than an instantaneous bottom-up signal; it adds no parameters and applies to existing diagonal SSMs as well as to RQFs.

The biological motivation for this correction comes from \textit{prospective} firing: under sinusoidally modulated somatic current injection, cortical pyramidal cells can fire at a phase that leads their input, and biophysical models explain this as a firing-rate advance relative to membrane voltage~\citep{kondgen2008dynamical,brandt2024prospective}. Apical dendritic stimulation can likewise modulate the soma with little integration lag through dendritic resonance~\citep{ulrich2002dendritic,brandt2024prospective}.
Related prospective responses have been reported in retinal ganglion-cell populations, cerebellar Purkinje cells, and human muscle-spindle afferents~\citep{palmer2015predictive,ostojic2015neuronal,dimitriou2010human}.
Such phase leads can be modeled by applying the \textit{prospective} (or look-ahead) operator~\citep{haider2021latent,senn2024neuronal,ellenberger2025backpropagation}
\begin{equation}
    \chit \;\equiv\; 1 + \tau\,\frac{\mathrm{d}}{\mathrm{d}t}
    \label{eq:chit_def}
\end{equation}
to the underlying voltage or firing rate. For a first-order low-pass filter, $\chit$ is its exact differential inverse, so applying it cancels the filter's lag. When interpreted as a time-advance operator, however, it is accurate only to first order: expanding the unit time-shift in $\tau$ gives $u(t+\tau) \;=\; u(t) + \tau\,\dot u(t) + O(\tau^2) \;=\; \chit(u(t)) + O(\tau^2)$. Thus, $\chit$ advances a smooth signal by one membrane time constant to first order in $\tau$.

We apply this prospective operator in networks of RQFs, a class of biologically motivated complex-valued temporal band-pass filters~\citep{heeger2020recurrent}. We derive RQFs from an energy-based theory of cortical function~\citep{heeger2017theory}, in which a state parameter tunes a compromise between feedforward sensory drive and an internally generated prior. Taking the prior to be the prospective image of the state under a phase-rotation model collapses the resulting dynamics into a scalar complex-valued band-pass filter with a learnable phase-prior frequency $\omega$ and a learnable bandwidth parameter $\gamma$ (Section~\ref{sec:rqf}).

It has been suggested that the brain relies on a set of canonical neural microcircuits, across brain regions and modalities, implementing hierarchies of spatial band-pass filtering, output nonlinearities, and spatial pooling that extract and combine features~\citep{douglas1995recurrent,heeger1996computational,simoncelli1998model,riesenhuber1999hierarchical,riesenhuber2002neural,carandini2012normalization,oleskiw2025foundations}, much as convolutional neural networks do. RQFs extend this paradigm to extract and combine features over time. Temporal band-pass RQFs have been used to model temporal-frequency selectivity and direction selectivity of neurons in primary visual cortex~\citep{heeger2020recurrent}. The encoding of sound in the cochlea and the auditory nerve can likewise be modeled as a bank of band-pass filters~\citep{von1960experiments}.

We prove that scalar RQFs are a special case of single-channel diagonal complex SSMs, so an RQF layer is a constrained diagonal SSM with biologically interpretable poles and inherits the parallel-scan and FFT-convolutional training algorithms used for SSMs~\citep{gu2022parameterization,gupta2022diagonal}. This distinguishes RQFs from recent prospective-coding cortical models: neuronal least-action~\citep{senn2024neuronal} (NLA) and latent equilibrium~\citep{haider2021latent} use prospective signals for local biological learning, while generalized latent equilibrium~\citep{ellenberger2025backpropagation} (GLE) uses prospective and retrospective operators for online credit assignment in a feedforward architecture. RQFs instead provide recurrent temporal filters compatible with long-sequence SSM training.

We show, both theoretically and empirically, that deep RQF networks improve when each layer's bottom-up input drive is prospective. Prospective and instantaneous-input RQFs share the same prospective phase-rotation prior via recurrence; they differ only in whether the bottom-up drive is look-ahead coded. This input-side change is local to each layer, leaves the RQF parameters and state transition unchanged, admits a two-tap discretization applicable to diagonal SSMs (Section~\ref{sec:method}), and directly affects credit assignment under \textit{spatial-only backpropagation}. Spatial-only backpropagation treats every earlier recurrent state as a stop-gradient variable, so it computes gradients only across layers, not over time. Under this regime, instantaneous inputs introduce an explicit $(h/\tau)^{L-\ell}$ attenuation (where $h$ is the discretization time step), while prospective-input coding changes the discretization prefactor of each spatial hop from $O(h/\tau)$ to $O(1)$ as $h/\tau\to 0$.

We also introduce \textit{selective RQFs} (sRQFs), an extension of RQFs with attention-like selection. Selective SSMs make the input matrix depend on the current input, so that the model controls which inputs enter its state~\citep{gu2023mamba}. One way to do this in an RQF layer is to modulate the input with an input-dependent gain: each sRQF filter multiplies its bottom-up drive by a positive gain computed from the preceding layer's output, analogous to the input gate of LSTMs and the input modulators of ORGaNICs~\citep{hochreiter1997long,heeger2019oscillatory}. Akin to Mamba, which also makes its discretization step input-dependent~\citep{gu2023mamba}, an sRQF can additionally make the bandwidth $\gamma$ and frequency $\omega$ parameters input-dependent, so that each filter's pole moves with the input. Because these quantities depend only on the preceding layer's output and not on the layer's own state, sRQFs remain diagonal linear recurrences, time-varying when $\gamma$ and $\omega$ are selective, that can be computed efficiently using a parallel scan~\citep{blelloch1990prefix,martin2018parallelizing,smith2022simplified,gu2023mamba}, with either prospective or instantaneous input (Appendix~\ref{app:selective_rqf}).

\noindent\textbf{Contributions.}
\begin{itemize}[leftmargin=*, itemsep=1pt]
    \item \textit{Architecture (machine learning).} We introduce RQFs, which are motivated by neuroscience but developed here for machine-learning applications. We prove that RQFs are a special case of diagonal SSMs (Section~\ref{sec:rqf}), making RQF layers compatible with the parallel-scan and FFT-convolution algorithms used for diagonal SSMs, thereby making RQF stacks capable of learning tasks such as the $16{,}384$-step Path-X. We also introduce sRQFs, which make the input gain, bandwidth $\gamma$, and frequency $\omega$ parameters input-dependent while retaining the parallel scan (Appendix~\ref{app:selective_rqf}).
    \item \textit{Prospective input (machine learning).} We replace each layer's instantaneous bottom-up drive with a prospective one and derive the corresponding discretization for any diagonal SSM (Section~\ref{sec:method}). The implementation adds no parameters, leaves the state transition and parallel scan untouched, and costs only a second input tap; under full BPTT, it yields higher mean test accuracy than the matched control in every RQF configuration.
    \item \textit{Theory (computational neuroscience and machine learning).} We prove that under spatial-only backpropagation, instantaneous bottom-up inputs incur a geometric discretization-induced depth attenuation that prospective-input coding removes (Section~\ref{sec:method} and Appendix~\ref{app:gradient_proof}).
    \item \textit{Performance (machine learning).} We evaluate prospective input on Speech Commands and Path-X with RQF stacks, S5, and the nonlinear ORGaNICs substrate~\citep{heeger2019oscillatory}, under full BPTT and spatial-only backpropagation, and compare RQFs with sRQFs on Path-X and Selective Copying (Section~\ref{sec:experiments}).
\end{itemize}

\noindent\textbf{Scope of contributions.} The RQF substrate and the input-side correction are evaluated as machine-learning contributions under standard BPTT; the cortical theory provides their motivation. The third contribution concerns spatial-only backpropagation, in which the recurrence is never differentiated through time. We adopt that regime because it is the one described by our derivation and isolates the effect of the input drive; it is a diagnostic, neither a prerequisite for the machine-learning contribution nor a recommended training procedure. Spatial-only backpropagation is temporally local, but exact error propagation across layers still uses conjugate-transposed forward weights and therefore retains the weight-transport problem~\citep{lillicrap2016random}. We consequently describe RQFs as a biologically motivated circuit abstraction but do not suggest that backpropagation training is biologically plausible.

\section{Recursive Quadrature Filters (RQFs)}
\label{sec:rqf}

\begin{figure}[t]
    \centering
    \includegraphics[width=\linewidth]{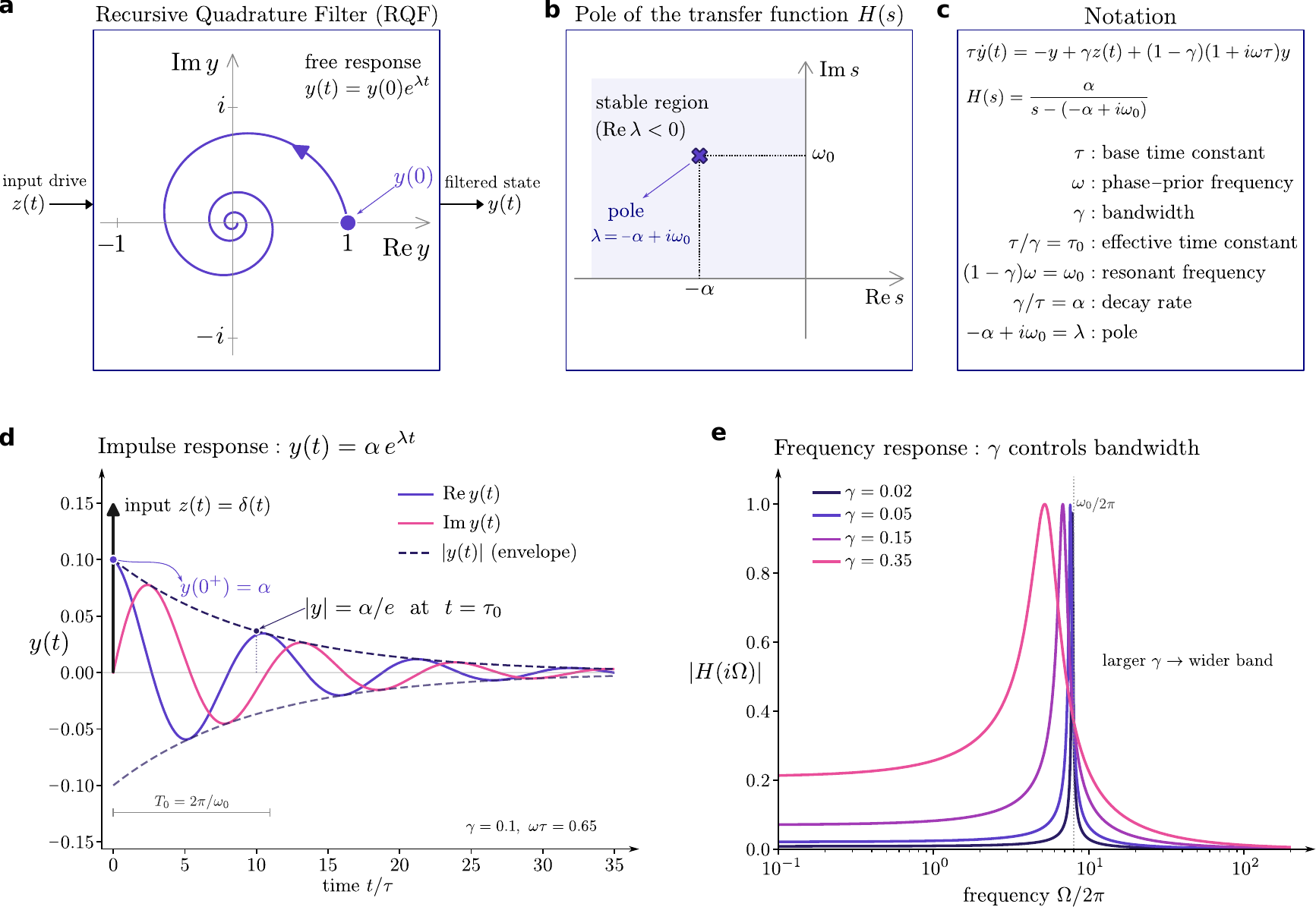}
    \caption[The Recursive Quadrature Filter: pole, impulse response, and frequency response.]{\textbf{The Recursive Quadrature Filter (RQF): pole, impulse response, and frequency response.} An RQF is a scalar complex-valued temporal filter with learnable phase-prior frequency $\omega$ and bandwidth parameter $\gamma$ at a fixed base time constant $\tau$, governed by Eq.~\eqref{eq:rqf_scalar}. The derived quantities are the decay rate $\alpha = \gamma/\tau$, resonant frequency $\omega_0 = (1-\gamma)\omega$, effective time constant $\tau_0 = \tau/\gamma$, and pole $\lambda = -\alpha + \iota\omega_0$. \textbf{(a)} Free response of $y$ in the complex plane, $y(t) = y(0)\,e^{\lambda t}$. \textbf{(b)} The pole lies in the stable region $\Real(\lambda) < 0$ for any $\gamma \in (0, 1)$. \textbf{(c)} Notation used throughout the paper. \textbf{(d)} Impulse response $y(t) = \alpha\,e^{\lambda t}$ to a Dirac drive $z(t) = \delta(t)$, a quadrature pair (i.e., with nearly equal amplitude responses and a nearly $90^\circ$ phase shift) oscillating inside the envelope $|y(t)| = \alpha\,e^{-t/\tau_0}$ ($\gamma = 0.1$, $\omega\tau = 0.65$). \textbf{(e)} Frequency response amplitude for several values of $\gamma$, from sharp resonance to broad passband.}
    \label{fig:figure1}
\end{figure}

The substrate for RQFs is a leaky integrator whose state relaxes toward a convex mixture of the external input and an internally predicted recurrent state:
\begin{equation}
    \tau\,\dot y \;=\; -y + \left[\gamma\,z + (1-\gamma)\,\widehat y\right],
    \label{eq:leaky_integrator}
\end{equation}
where $y(t) \in \C$ is the filter state; $z(t) \in \C$ is the input drive; $\tau > 0$ is the base time constant; $\widehat y$ is the recurrent prediction; and $\gamma \in (0,1)$ controls the balance between input tracking and recurrence. This convex blend mirrors the cortical energy model of~\citet{heeger2017theory}; Appendix~\ref{app:energy_derivation} derives Eq.~\eqref{eq:leaky_integrator} from that energy functional, with $\gamma$ as the state parameter and $\widehat y$ as the prior drive.

The recurrent prediction $\widehat y$ in Eq.~\eqref{eq:leaky_integrator} requires a model of how $y$ evolves locally. We take this model to be a \textit{phase-rotation prior}: the undamped complex oscillator $\dot y = \iota\omega y$, the minimal generator on $\C$ of a single phase-prior frequency $\omega$. This choice follows the internal-model principle: a system that prospectively tracks a structured signal should contain a model of that signal's dynamics~\citep{francis1976internal}. Applying the prospective operator $\chit$ from Eq.~\eqref{eq:chit_def} to this prior gives
\begin{equation}
    \widehat y \;\triangleq\; \chit(y)\big|_{\dot y = \iota\omega y} \;=\; (1+\iota\omega\tau)\,y,
    \label{eq:yhat_def}
\end{equation}
where $\chit(y(t))$ is causal because it uses only the instantaneous activity and its local time derivative.

Substituting Eq.~\eqref{eq:yhat_def} into Eq.~\eqref{eq:leaky_integrator} yields the \textit{Recursive Quadrature Filter} (RQF), a scalar complex-valued temporal band-pass filter governed by
\begin{equation}
    \tau\,\dot y \;=\; -y + \gamma\,z + (1-\gamma)(1 + \iota\omega\tau)\,y,
    \label{eq:rqf_scalar}
\end{equation}
where $\omega > 0$ is the learnable frequency parameter and $\gamma$ is the learnable bandwidth parameter. The state $y$ is complex-valued even when $z$ is real, with its real and imaginary parts representing the outputs of a quadrature pair of causal temporal band-pass filters (i.e., for narrow passbands, nearly equal amplitude responses and a nearly $90^\circ$ phase shift). 
Hence, the modulus of the output $|y|$
follows the amplitude envelope of the filter's band
regardless of phase.
Defining $\alpha \triangleq \gamma/\tau$, $\omega_0 \triangleq (1-\gamma)\omega$, and $\lambda \triangleq -\alpha + \iota\,\omega_0$ collapses Eq.~\eqref{eq:rqf_scalar} into the compact scalar recurrence $\dot y = \lambda y + \alpha z$, so the RQF is a stable single-pole complex filter for any $\gamma \in (0,1)$ (Fig.~\ref{fig:figure1}). The prospective factor $(1+\iota\omega\tau)y$ sets the local recurrent prediction, whereas the homogeneous state evolves as $y(t+\Delta)=e^{\lambda\Delta}y(t)$ when $z=0$. Appendix~\ref{app:linear_systems} derives the transfer function (revealing the band-pass property), quality factor, damped-oscillator form, and the effect of cascading RQFs for fine-tuning the bandwidth.

\noindent\textbf{Multi-layer architecture.} A layer of an RQF stack bundles $n_\ell$ scalar filters operating in parallel, each with its own phase-prior frequency $\omega_i^{(\ell)}$ and bandwidth $\gamma_i^{(\ell)}$ but sharing a common time constant $\tau$. Collecting the parameters into vectors $\bomega^{(\ell)}, \bgamma^{(\ell)} \in \R^{n_\ell}$ and writing the layer state as $\by^{(\ell)} \in \C^{n_\ell}$, the layer dynamics are the element-wise vectorization of Eq.~\eqref{eq:rqf_scalar}:
\begin{equation}
    \tau\,\dot{\by}^{(\ell)} \;=\; -\by^{(\ell)} \;+\; \bgamma^{(\ell)} \odot \bz^{(\ell)} \;+\; (\bone - \bgamma^{(\ell)}) \odot (\bone + \iota\tau\,\bomega^{(\ell)}) \odot \by^{(\ell)},
    \label{eq:rqf_layer_full}
\end{equation}
where $\bz^{(\ell)} \in \C^{n_\ell}$ is the bottom-up input drive to the layer; $\odot$ denotes the element-wise product; and $\bone \in \R^{n_\ell}$ is the all-ones vector. Defining $\balpha^{(\ell)} \triangleq \bgamma^{(\ell)}/\tau$ and $\blambda^{(\ell)} \triangleq -\balpha^{(\ell)} + \iota\,(\bone-\bgamma^{(\ell)})\odot\bomega^{(\ell)}$, Eq.~\eqref{eq:rqf_layer_full} collapses to the compact diagonal complex ODE:
\begin{equation}
    \dot{\by}^{(\ell)} \;=\; \blambda^{(\ell)} \odot \by^{(\ell)} \;+\; \balpha^{(\ell)} \odot \bz^{(\ell)}.
    \label{eq:rqf_layer}
\end{equation}

To build a deep network, we stack $L$ layers. The input drive to layer $\ell$ is
\begin{equation}
    \bz^{(\ell)} \;=\; \bW^{(\ell)} \bx^{(\ell)}, \qquad \bx^{(\ell)} \;=\; \rho\!\left(\by^{(\ell-1)}\right),
    \label{eq:layer_drive}
\end{equation}
where $\rho:\C^{n_{\ell-1}} \to \C^{n_{\ell-1}}$ is an activation function (also called an output nonlinearity); $\bx^{(\ell)} \in \C^{n_{\ell-1}}$ is the \textit{layer input} to layer $\ell$, with the boundary $\bx^{(1)} \triangleq \br_{\mathrm{in}}$; $\bW^{(\ell)} \in \C^{n_\ell \times n_{\ell-1}}$ is a complex-valued inter-layer weight matrix; and $\bz^{(\ell)} \in \C^{n_\ell}$ is the post-weight bottom-up input drive. The activation function $\rho$ is not required to be holomorphic: nothing in the layer dynamics or in the discretization of Section~\ref{sec:method} relies on complex differentiability, and Wirtinger calculus supplies gradients for any real-differentiable map of $(\Real(\by), \Imag(\by))$. The default is the \textit{pointwise} family $[\rho(\by)]_i = \rho_0(y_i)$, with the canonical choice being the split-ReLU
\begin{equation}
    \rho_0(y) \;=\; \max\{0,\Real(y)\} \;+\; \iota\,\max\{0,\Imag(y)\},
    \label{eq:split_relu}
\end{equation}
which applies a real-valued rectifier independently to the real and imaginary components of the quadrature pair, and is non-holomorphic by construction.

\noindent\textbf{Discretization.} For the implementation, we discretize the continuous-time layer dynamics in Eq.~\eqref{eq:rqf_layer} with a \textit{zero-order hold} (ZOH) on the bottom-up drive $\bz^{(\ell)}$. ZOH treats $\bz^{(\ell)}$ as piecewise constant on each integration interval $[t_n, t_{n+1}]$, with the constant value denoted by $\bz_{n+1}^{(\ell)}$, and integrates the resulting forced linear ODE in closed form. It is the default discretization for diagonal SSMs, adopted by DSS, S4D, S5, and Mamba~\citep{gupta2022diagonal,gu2022parameterization,smith2022simplified,gu2023mamba}, because it (i) is unconditionally stable for any pole in the open left half-plane, meaning that high-$Q$ (narrow passband) filters impose no extra step-size constraint; (ii) integrates the linear part exactly, retaining the full $e^{\blambda h}$ response at finite $h$; and (iii) agrees with forward Euler to first order in $h/\tau$, with an $O((h/\tau)^2)$ difference.
ZOH discretization of Eq.~\eqref{eq:rqf_layer} yields the diagonal complex affine recurrence and coefficients
\begin{align}
    \by_{n+1}^{(\ell)} \;&=\; \bar{\bA}_h^{(\ell)} \odot \by_n^{(\ell)} \;+\; \bar{\bB}_h^{(\ell)} \odot \bz_{n+1}^{(\ell)},
    \label{eq:zoh_main}\\
    \bar{\bA}_h^{(\ell)} \;&\triangleq\; e^{\blambda^{(\ell)} h}, \qquad
    \bar{\bB}_h^{(\ell)} \;\triangleq\; \frac{\balpha^{(\ell)}}{\blambda^{(\ell)}} \odot \!\left(\bar{\bA}_h^{(\ell)} - \bone\right),
    \label{eq:AB_main}
\end{align}
where $\bz_{n+1}^{(\ell)} = \bW^{(\ell)} \rho(\by_{n+1}^{(\ell-1)})$ samples the bottom-up input drive in Eq.~\eqref{eq:layer_drive} at $t_{n+1}$. The coefficients $\bar{\bA}_h^{(\ell)}$ and $\bar{\bB}_h^{(\ell)}$ depend only on the parameters $(\bgamma^{(\ell)}, \bomega^{(\ell)})$, the base time constant $\tau$, and the step size $h$, and can be precomputed once per parameter update and reused across all time steps. The full derivation is given in Appendix~\ref{app:zoh}.

\begin{figure}[t]
    \centering
    \includegraphics[width=0.82\linewidth]{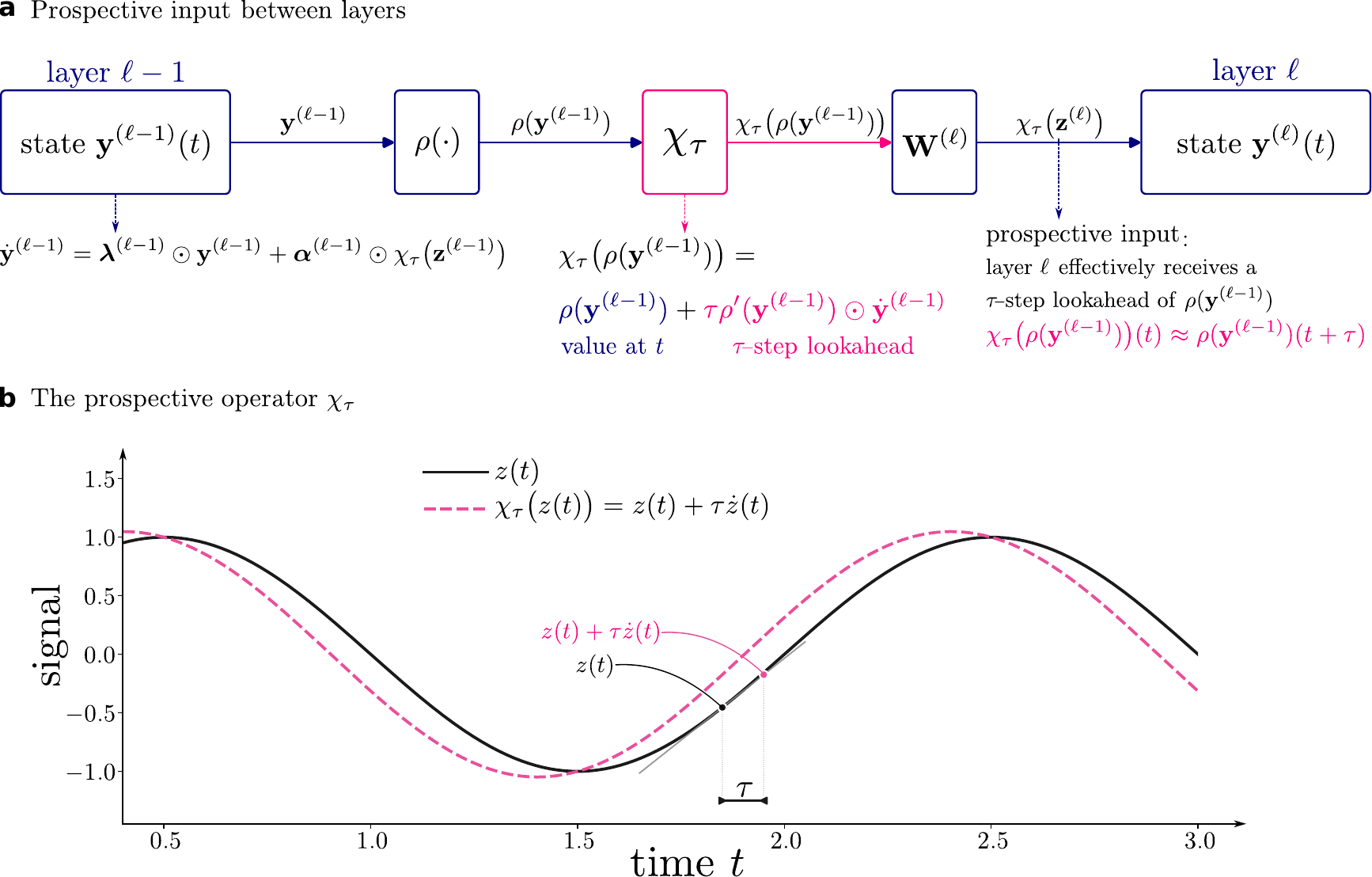}
    \caption[Prospective-input coding in multi-layer RQFs.]{\textbf{Prospective-input coding in multi-layer RQFs.} The \textit{prospective-input} variant replaces the bottom-up post-nonlinearity signal entering layer $\ell$ with its $\tau$-step look-ahead $\chit(\rho(\by^{(\ell-1)}))$ defined in Eq.~\eqref{eq:prospective_input}; the \textit{instantaneous-input} baseline feeds $\rho(\by^{(\ell-1)})$ directly into $\bW^{(\ell)}$. \textbf{(a)} The state $\by^{(\ell-1)}$ is passed through the activation $\rho$, advanced by the prospective operator $\chit$, and mixed by the inter-layer weight matrix $\bW^{(\ell)}$ to form the bottom-up drive $\chit(\bz^{(\ell)})$. \textbf{(b)} Action of $\chit$ on a sinusoidal signal $z(t)$ (black). The look-ahead $\chit(z(t)) = z(t) + \tau\,\dot{z}(t)$ (pink, dashed) advances $z$ by approximately $\tau$ by extrapolating along the local tangent.}
    \label{fig:figure2}
\end{figure}

\noindent\textbf{RQFs are a special case of diagonal state-space models.} A continuous-time linear state-space model (SSM) maps an input signal $\bx(t) \in \R^{U}$ through a latent state $\by(t) \in \R^{P}$ to an output $\bo(t) \in \R^{M}$ via
\begin{equation}
    \dot{\by}(t) \;=\; \bA\,\by(t) + \bB\,\bx(t), \qquad \bo(t) \;=\; \bC\,\by(t) + \bD\,\bx(t),
    \label{eq:ssm_continuous}
\end{equation}
with state matrix $\bA \in \R^{P\times P}$, input matrix $\bB \in \R^{P\times U}$, output matrix $\bC \in \R^{M\times P}$, and feedthrough $\bD \in \R^{M\times U}$~\citep{gu2021efficiently,smith2022simplified}. All SSMs share this template but differ in how the recurrent core is parameterized and how the feature channels are coupled. S4 applies a bank of single-input single-output (SISO) SSMs across feature channels~\citep{gu2021efficiently}; DSS and S4D replace S4's structured state matrix with diagonal variants~\citep{gupta2022diagonal,gu2022parameterization}; S5 uses one multi-input multi-output (MIMO) SSM whose effective recurrent matrix is complex-valued and diagonal~\citep{smith2022simplified}; the LRU uses a discrete diagonal linear recurrence with dense input and output projections~\citep{orvieto2023resurrecting}; and Mamba retains a diagonal state transition while making the discretized parameters input-dependent~\citep{gu2023mamba}. In the fixed-parameter diagonal case, the recurrent core can be written directly in a complex diagonal form,
\begin{equation}
    \dot{\widetilde{\by}}(t) \;=\; \bLambda\,\widetilde{\by}(t) + \bB\,\bx(t), \qquad \bo(t) \;=\; \bC\,\widetilde{\by}(t) + \bD\,\bx(t),
    \label{eq:ssm_diagonal_core}
\end{equation}
with $\bLambda \in \C^{P\times P}$ diagonal, $\bB \in \C^{P\times U}$, and $\bC \in \C^{M\times P}$. ZOH discretization at step $h$ gives $\widetilde{\by}_{k+1} = \bar{\bA}_h\,\widetilde{\by}_k + \bar{\bB}_h\,\bx_{k+1}$, with
\begin{equation}
    \bar{\bA}_h \;\triangleq\; e^{\bLambda h}, \qquad
    \bar{\bB}_h \;\triangleq\; \bLambda^{-1}\!\left(\bar{\bA}_h-\bI\right)\bB.
    \label{eq:ssm_zoh_bar}
\end{equation}
The temporal modes are controlled by the entries of $\bLambda$: an eigenvalue $\lambda_p \in \C$ in the open left half-plane produces a damped oscillation at angular frequency $\Imag(\lambda_p)$ with decay rate $|\Real(\lambda_p)|$.

The RQF layer is the following constrained special case of Eq.~\eqref{eq:ssm_diagonal_core}, obtained by substituting the layer drive from Eq.~\eqref{eq:layer_drive} into Eq.~\eqref{eq:rqf_layer}:
\begin{equation}
    \dot{\by}^{(\ell)}
    \;=\;
    \mathrm{diag}(\blambda^{(\ell)})\,\by^{(\ell)}
    +
    \mathrm{diag}(\balpha^{(\ell)})\,\bW^{(\ell)}\bx^{(\ell)}.
    \label{eq:rqf_as_ssm}
\end{equation}
Thus, the SSM input $\bx(t)$ corresponds to the RQF layer input $\bx^{(\ell)}$, and the dense input projection is the inter-layer weight matrix $\bW^{(\ell)}$. In the canonical SSM notation of Eq.~\eqref{eq:ssm_diagonal_core}, the full continuous-time input matrix is $\bB = \mathrm{diag}(\balpha^{(\ell)})\bW^{(\ell)}$; the diagonal factor is not an extra SSM parameter but the RQF input gain already present in Eq.~\eqref{eq:rqf_layer}. The RQF-specific constraint is that $\balpha^{(\ell)} = \bgamma^{(\ell)}/\tau$ and $\blambda^{(\ell)} = -\balpha^{(\ell)} + \iota(\bone-\bgamma^{(\ell)})\odot\bomega^{(\ell)}$, which identifies the channel input gain with the inverse decay timescale $1/\tau_0^{(\ell)}$. The inclusion is one-way: a general diagonal SSM need not satisfy this pole--input-gain coupling and therefore need not be an RQF. The remaining SSM machinery transfers without modification: any output projection $\bC \in \C^{M\times n_\ell}$ and feedthrough $\bD \in \C^{M\times U}$ may be appended to read out from $\by^{(\ell)}$, and the per-block residual connections used in S4, S5, and Mamba stacks are admissible at the layer level. Unless stated otherwise, we set $\bC = \bI$ and $\bD = \bzero$ throughout, so each layer's output is its own complex state $\by^{(\ell)}$ and inter-layer mixing is delegated to $\bW^{(\ell+1)}$; residual connections are treated as an explicit configuration choice in our experiments (Section~\ref{sec:experiments}).

\noindent\textbf{Selective RQFs (sRQFs).} An sRQF multiplies each filter's bottom-up drive by a positive gain $\bm{\kappa}^{(\ell)}$ computed from the preceding layer's output, so that the input matrix of Eq.~\eqref{eq:rqf_as_ssm} becomes input-dependent:
\begin{equation}
    \bB_{\kappa}^{(\ell)} = \mathrm{diag}(\bm{\kappa}^{(\ell)})\,\mathrm{diag}(\balpha^{(\ell)})\,\bW^{(\ell)}.
    \label{eq:srqf}
\end{equation}
An sRQF can additionally make the bandwidth $\bgamma^{(\ell)}$ and frequency $\bomega^{(\ell)}$ parameters input-dependent, so that each filter's pole, and hence its state transition, varies with the input, as Mamba's state transition does through its input-dependent discretization step~\citep{gu2023mamba}. For the results reported below, only the input gain was input-dependent for the Path-X benchmark, whereas the gain, bandwidth $\bgamma$, and frequency $\bomega$ were all input-dependent for the Selective Copying benchmark (Section~\ref{sec:experiments}). Appendix~\ref{app:selective_rqf} gives the parameterization, the discretized updates, and their parallel scan.

\section{Prospective-input coding in multi-layer RQFs}
\label{sec:method}

We use the terms \textit{instantaneous-input} and \textit{prospective-input} to refer only to the bottom-up input drive from layer $\ell-1$ to layer $\ell$. Both variants retain the RQF recurrent prediction $\widehat y=(1+\iota\omega\tau)y$ from Section~\ref{sec:rqf}; the comparison changes the input path, not the RQF dynamics. With the notation of Eq.~\eqref{eq:layer_drive}, the instantaneous choice uses $\bx^{(\ell)}=\rho(\by^{(\ell-1)})$, while the prospective-input alternative applies the look-ahead operator before the inter-layer weights act (Fig.~\ref{fig:figure2}):
\begin{equation}
    \widetilde{\bx}^{(\ell)}
    \;\triangleq\;
    \chit\!\left(\bx^{(\ell)}\right)
    \;=\;
    \rho\!\left(\by^{(\ell-1)}\right) + \tau\,\dot{\rho}\!\left(\by^{(\ell-1)}\right),
    \qquad
    \bz_\chi^{(\ell)} \;=\; \bW^{(\ell)}\widetilde{\bx}^{(\ell)},
    \label{eq:prospective_input}
\end{equation}
where the dot denotes the time derivative. Since $\bW^{(\ell)}$ is fixed over the integration interval, applying $\chit$ to $\bx^{(\ell)}$ and then multiplying by $\bW^{(\ell)}$ is equivalent in the forward pass to applying $\chit$ to $\bz^{(\ell)}$; the discretization below uses this compact input drive notation.

\noindent\textbf{Discretization.} Applying the ZOH update to the prospective input drive $\chit\left(\bz^{(\ell)}\right)$ gives the following recurrence; Appendix~\ref{app:impulse_zoh} provides the derivation:
\begin{equation}
    \by_{n+1}^{(\ell)} \;=\; \bar{\bA}_h^{(\ell)} \odot \by_n^{(\ell)} \;+\; \left[\bar{\bB}_h^{(\ell)} + \bgamma^{(\ell)} \odot \bar{\bA}_h^{(\ell)}\right] \odot \bz_{n+1}^{(\ell)} \;-\; \bgamma^{(\ell)} \odot \bar{\bA}_h^{(\ell)} \odot \bz_n^{(\ell)},
    \label{eq:zoh_prospective_main}
\end{equation}
where $\bz_k^{(\ell)} = \bW^{(\ell)}\bx_k^{(\ell)}$ and $\bx_k^{(\ell)} = \rho(\by_k^{(\ell-1)})$. In standard signal-processing terms, Eq.~\eqref{eq:zoh_prospective_main} is a \textit{first-order IIR filter with a two-tap feedforward path}: the recurrence on $\by^{(\ell)}$ has a single feedback pole $\bar{\bA}_h^{(\ell)}$, while the compact drive enters through weighted taps at $\bz_{n+1}^{(\ell)}$ and $\bz_n^{(\ell)}$.

Prospective input adds no parameters and leaves the diagonal state transition and associative scan unchanged. For a length-$T$ sequence, it adds only $O(Tn_\ell)$ element-wise work and one lagged read of the drive, so the asymptotic cost of inference and full BPTT is unchanged (Appendix~\ref{app:impulse_zoh}).

\noindent\textbf{Why prospective-input coding improves the spatial gradient path.} Spatial-only backpropagation freezes the temporal recurrence at the previous time step and leaves only the within-step chain through depth. An instantaneous-input Euler step sends each inter-layer error through $\mathbf{g}_{\mathrm{np}}=(h/\tau)\bgamma$, whereas prospective input uses $\mathbf{g}_{+}=(1+h/\tau)\bgamma$. As $h/\tau\to 0$, therefore, the discretization prefactor multiplying each spatial hop changes from $O(h/\tau)$ to $O(1)$ and the explicit $(h/\tau)^{L-\ell}$ factor disappears. This statement concerns that prefactor only: the common product of activation slopes, weight norms, and damping gains may still attenuate with depth, and the upper bounds in Appendix~\ref{app:gradient_proof} do not guarantee an $O(1)$ complete gradient. The implemented ZOH update has the same small-$h$ leading-order coefficients. Spatial-only backpropagation is local in time but still uses $\bW^\dagger$ to propagate exact adjacent-layer errors, so it retains the weight-transport problem and is used here only to isolate the analyzed path.

\section[Experiments]{Experiments\protect\footnote{Code for the experiments is available at \url{https://github.com/Sequel-Institute/prospective-rqf}.}}
\label{sec:experiments}

RQFs are band-pass filters by construction (Appendix~\ref{app:linear_systems}), so speech provides a natural first test of whether prospective-input coding improves learning through the RQF cascade. We use the 10-class subset of Google Speech Commands v0.02~\citep{warden2018speech}, comprising one-second utterances of the keywords \textit{yes}, \textit{no}, \textit{up}, \textit{down}, \textit{left}, \textit{right}, \textit{on}, \textit{off}, \textit{stop}, and \textit{go}. We consider two representations of each utterance: the raw waveform, supplied as a sequence of $16{,}000$ scalar samples, and $20$ Mel-frequency cepstral coefficients (MFCCs), supplied as a sequence of $161$ frames. Thus, the two representations pose the same classification problem at very different sequence lengths and levels of temporal abstraction.

We contrast two regimes for assigning gradient credit through the recurrence. Full backpropagation through time (BPTT) differentiates through the complete recurrent trajectory. Spatial-only backpropagation instead detaches every previous recurrent state and retains only the within-step chain through depth. We use the latter as a mechanistic diagnostic, not as the recommended optimizer, because it removes temporal gradient credit and exposes the spatial attenuation analyzed in Section~\ref{sec:method}.

\begin{table}[t!]
    \centering
    \caption{\textbf{RQF on Speech Commands.} Test accuracy for non-residual width-64 stacks trained with \textbf{(a)} full BPTT and \textbf{(b)} spatial-only backpropagation.}
    \label{tab:rqf_bptt}
    \label{tab:rqf_spatial}
    \setlength{\tabcolsep}{4pt}
    \renewcommand{\arraystretch}{0.95}
    \begin{tabular}{c c cc c cc}
        \toprule
        & \multicolumn{3}{c}{Raw audio} & \multicolumn{3}{c}{MFCC} \\
        \cmidrule(lr){2-4}\cmidrule(lr){5-7}
        Depth & Params & Prospective & Non-pros. & Params & Prospective & Non-pros. \\
        \midrule
        \multicolumn{7}{l}{\textbf{(a) Full BPTT}} \\
        2 & $52.3$k & $94.94\!\pm\!0.10$ & $94.40\!\pm\!0.17$ & $53.5$k & $96.42\!\pm\!0.26$ & $96.30\!\pm\!0.33$ \\
        4 & $68.9$k & $96.27\!\pm\!0.16$ & $95.77\!\pm\!0.22$ & $70.2$k & $96.65\!\pm\!0.24$ & $96.41\!\pm\!0.14$ \\
        6 & $85.6$k & $96.35\!\pm\!0.20$ & $96.13\!\pm\!0.12$ & $86.8$k & $96.70\!\pm\!0.19$ & $96.49\!\pm\!0.07$ \\
        \addlinespace[2pt]
        \multicolumn{7}{l}{\textbf{(b) Spatial-only backpropagation}} \\
        2 & $52.3$k & $67.76\!\pm\!7.19$ & $78.11\!\pm\!0.71$ & $53.5$k & $94.89\!\pm\!0.20$ & $94.20\!\pm\!0.28$ \\
        4 & $68.9$k & $84.70\!\pm\!3.17$ & $81.70\!\pm\!1.96$ & $70.2$k & $92.63\!\pm\!0.85$ & $89.04\!\pm\!0.83$ \\
        6 & $85.6$k & $82.92\!\pm\!5.42$ & $75.34\!\pm\!1.18$ & $86.8$k & $91.55\!\pm\!0.76$ & $76.38\!\pm\!1.57$ \\
        \bottomrule
    \end{tabular}
\end{table}

The cleanest test is the RQF stack itself. We therefore begin with non-residual RQF models of depth $L\in\{2,4,6\}$ and width $n\in\{32,64\}$, using split-ReLU between layers and no normalization or gating. Avoiding residual connections leaves no bypass around the recurrent substrate, so changes in spatial credit flow are attributed to the input drive. Within every RQF comparison, the prospective and non-prospective variants match in architecture, optimizer, schedule, seeds, and parameter count; the look-ahead scale and base time constant are both $\tau=5h$. All tables report test accuracy as the mean $\pm$ sample standard deviation over five seeds (for Selective Copying, over the seeds that exceed chance performance), and Appendices~\ref{app:hyperparams} and~\ref{app:selective_copying_protocol} give the parameterization and training protocol.

Under full BPTT, prospective-input coding raises mean test accuracy in all matched depth, width, and feature comparisons (Tables~\ref{tab:rqf_bptt} and~\ref{tab:rqf_bptt_width32}). Adding residual connections between layers leaves these results unchanged: prospective input still gives the higher mean in every configuration (Appendix~\ref{app:residual_results}, Table~\ref{tab:rqf_bptt_residual}). The smallest raw-audio models achieve $94.59\pm0.28\%$, $95.66\pm0.23\%$, and $96.09\pm0.23\%$ at two, four, and six layers, respectively, while using only $23.5$k--$31.9$k parameters (Table~\ref{tab:rqf_bptt_width32}). The spatial-only runs reveal the predicted depth dependence more clearly (Table~\ref{tab:rqf_spatial}): on MFCC inputs, the prospective advantage grows from $0.69$ to $3.59$ to $15.17$ points as depth increases.

The direct gradient measurements connect this behavioral result to the analysis. Under spatial-only backpropagation, prospective inputs produce larger spatial per-hop gains and early-layer weight-gradient norms than instantaneous inputs. Thus, the advantage of prospective input is evident when learning depends on the spatial path isolated by the theory (Appendix~\ref{app:gradient_measurements}; Tables~\ref{tab:gradient_norms} and~\ref{tab:gradient_sweep}).

We next test whether the prospective construction transfers to a conventional residual SSM by comparing $\alpha$P-S5 with native S5~\citep{smith2022simplified}. We define $\alpha$P-S5 by scaling each row of the learned S5 input matrix with the corresponding native-clock pole decay rate $\alpha_p=-\Real(\lambda_p)$, i.e., $\bB_c=\mathrm{diag}(\balpha)\widetilde{\bB}$. Prospective input uses the shared physical look-ahead horizon $\tau=5h$, matching the RQF models. Appendix~\ref{app:s5_architecture} and Eq.~\eqref{eq:alpha_p_s5_taps} give the complete architecture and discretization. The mean test accuracy of $\alpha$P-S5 is higher at every tested depth, width, and input type (Tables~\ref{tab:s5_bptt} and~\ref{tab:s5_bptt_width32}).
These experiments place the minimal RQF architecture in context. At width $32$, the RQF is more accurate than $\alpha$P-S5 at every raw-audio depth while using fewer parameters; at six layers, it achieves $96.09\pm0.23\%$ with $31.9$k parameters, versus $95.06\pm0.99\%$ with $40.9$k for $\alpha$P-S5 (Tables~\ref{tab:rqf_bptt_width32} and~\ref{tab:s5_bptt_width32}).

\begin{table}[t!]
    \centering
    \caption{\textbf{$\alpha$P-S5 versus native S5 on Speech Commands, full BPTT.} Residual width-64 S5 blocks.}
    \label{tab:s5_bptt}
    \setlength{\tabcolsep}{3.2pt}
    \renewcommand{\arraystretch}{0.95}
    \begin{tabular}{c c cc c cc}
        \toprule
        & \multicolumn{3}{c}{Raw audio} & \multicolumn{3}{c}{MFCC} \\
        \cmidrule(lr){2-4}\cmidrule(lr){5-7}
        Depth & Params & $\alpha$P-S5 & Native S5 & Params & $\alpha$P-S5 & Native S5 \\
        \midrule
        2 & $48.9$k & $95.55\!\pm\!0.39$ & $94.94\!\pm\!0.42$ & $50.1$k & $96.52\!\pm\!0.20$ & $95.89\!\pm\!0.38$ \\
        4 & $74.2$k & $96.06\!\pm\!0.35$ & $95.96\!\pm\!0.27$ & $75.4$k & $96.40\!\pm\!0.17$ & $96.31\!\pm\!0.25$ \\
        6 & $99.5$k & $96.37\!\pm\!0.31$ & $96.21\!\pm\!0.23$ & $100.7$k & $96.41\!\pm\!0.20$ & $96.36\!\pm\!0.14$ \\
        \bottomrule
    \end{tabular}
\end{table}

To determine whether the effect depends on a linear RQF or SSM recurrence, we also apply the input correction to discrete residual ORGaNICs, a nonlinear recurrent substrate with dynamic, recurrent normalization~\citep{heeger2019oscillatory,rawat2024unconditional}. Prospective-input coding with full BPTT improves performance at two and four layers and ties at six layers. Under spatial-only backpropagation, the prospective model improves at four and six layers (Table~\ref{tab:organics_results}). The correction therefore transfers beyond the linear RQF core.

\begin{table}[t!]
    \centering
    \caption{\textbf{Discrete ORGaNICs on MFCC Speech Commands.} Residual width-64 models.}
    \label{tab:organics_results}
    \setlength{\tabcolsep}{4pt}
    \renewcommand{\arraystretch}{0.95}
    \begin{tabular}{c c cc cc}
        \toprule
        & & \multicolumn{2}{c}{Full BPTT} & \multicolumn{2}{c}{Spatial-only} \\
        \cmidrule(lr){3-4}\cmidrule(lr){5-6}
        Depth & Params & Prospective & Non-pros. & Prospective & Non-pros. \\
        \midrule
        2 & $58.3$k  & $93.28\!\pm\!0.37$ & $92.10\!\pm\!0.21$ & $79.88\!\pm\!1.28$ & $81.49\!\pm\!1.46$ \\
        4 & $124.3$k & $93.21\!\pm\!0.59$ & $91.91\!\pm\!0.53$ & $82.61\!\pm\!1.21$ & $81.67\!\pm\!0.88$ \\
        6 & $190.3$k & $91.92\!\pm\!0.19$ & $91.92\!\pm\!0.11$ & $83.43\!\pm\!1.73$ & $80.56\!\pm\!1.41$ \\
        \bottomrule
    \end{tabular}
\end{table}

\begin{table}[t!]
    \centering
    \begin{minipage}[b]{0.40\textwidth}
        \centering
        \caption{\textbf{Path-X, full BPTT.} Six-layer, width-64 models on the $16{,}384$-step task (chance: $50\%$). The sRQF uses only the input gain.}
        \label{tab:pathx_results}
        \setlength{\tabcolsep}{3.5pt}
        \renewcommand{\arraystretch}{0.95}
        \begin{tabular}{l c c}
            \toprule
            Model & Prospective & Non-pros. \\
            \midrule
            RQF & $83.36\!\pm\!1.71$ & $81.25\!\pm\!0.91$ \\
            sRQF & $86.31\!\pm\!0.51$ & $85.13\!\pm\!0.29$ \\
            \bottomrule
        \end{tabular}
    \end{minipage}\hfill
    \begin{minipage}[b]{0.57\textwidth}
        \centering
        \caption{\textbf{Selective Copying, full BPTT.} Four-layer, width-32 models (chance: $7.1\%$). The sRQF makes $\gamma$, $\omega$, and the input gain input-dependent. Parentheses give the number of seeds, out of five, that exceed chance performance.}
        \label{tab:selective_copying}
        \setlength{\tabcolsep}{3.5pt}
        \renewcommand{\arraystretch}{0.95}
        \begin{tabular}{l c c c}
            \toprule
            Model & Params & Prospective & Non-pros. \\
            \midrule
            RQF & $17.8$k & $69.83\!\pm\!1.78$\,(4/5) & $68.98\!\pm\!2.21$\,(2/5) \\
            sRQF & $33.0$k & $99.83\!\pm\!0.07$\,(5/5) & $99.72\!\pm\!0.23$\,(2/5) \\
            \bottomrule
        \end{tabular}
    \end{minipage}
    \vspace{-6pt}
\end{table}

Speech is well matched to a bank of temporal band-pass filters, but it does not establish whether RQFs with prospective input can learn structure that is neither acoustic nor naturally localized in frequency. We therefore turn to Path-X from Long Range Arena~\citep{tay2021long}, a binary classification task formed by flattening a $128\times128$ image into a $16{,}384$-step sequence whose relevant endpoints can be separated across the full input. We retain the forward-only, non-residual RQF model, use six layers of width $64$, and train with full BPTT using a Path-X-specific optimization schedule (Appendix~\ref{app:optimizer_training}).
The prospective and non-prospective models achieve $83.36\pm1.71\%$ and $81.25\pm0.91\%$, respectively (Table~\ref{tab:pathx_results}). Therefore, without bidirectionality, normalization, gating, or an expanded state, the minimal RQF stack learns a fully non-local $16{,}384$-step task more than $30$ points above chance. Prospective input yields better performance. When only the input gain is selective ($\gamma$ and $\omega$ fixed), the sRQF improves further to $86.31\pm0.51\%$ with prospective input and $85.13\pm0.29\%$ without it.

Finally, we compare RQFs and sRQFs on Selective Copying~\citep{gu2023mamba}, in which 16 data tokens are placed at random positions in a $4{,}096$-step sequence of blank tokens and must be reproduced in order at 16 final recall positions. The task requires content-aware selection of the tokens to be memorized~\citep{gu2023mamba}. \citet{gu2023mamba} argue that gating the input cannot affect the random spacing between these tokens; on this task, the sRQF therefore makes $\gamma$, $\omega$, and the input gain input-dependent (Eq.~\eqref{eq:srqf} and Appendix~\ref{app:selective_rqf}). We train four-layer, width-32 RQFs and sRQFs with both input rules for 600 epochs of 1,000 updates using full BPTT, supervise only the recall positions, and use five seeds per configuration. Unlike the Speech Commands and Path-X models, these models apply divisive normalization across filters after the split-ReLU (Appendix~\ref{app:selective_copying_protocol}). With prospective input, the sRQF solves the task in all five seeds, achieving $99.83\pm0.07\%$ test recall-token accuracy (Table~\ref{tab:selective_copying}). Without prospective input, only two seeds exceed chance ($99.72\pm0.23\%$), and the other three remain at chance. Prospective input likewise raises the number of RQF seeds that exceed chance, from two to four of five.

\vspace{-6pt}
\section{Related work}
\label{sec:related}

\noindent\textbf{Prospective coding in recurrent neural circuits.} Prospective coding has been used to address timing mismatch and credit assignment in biologically motivated learning rules. Latent equilibrium uses the look-ahead variable $u+\tau\dot u$ to recover fast feedforward computation and local error propagation~\citep{haider2021latent}. Neuronal least-action formulates a least-action principle over future-discounted voltages, yielding prospective firing rates and prospective somato-dendritic mismatch errors~\citep{senn2024neuronal}. Generalized latent equilibrium separates retrospective membrane integration from prospective output dynamics, yielding local spatio-temporal credit assignment~\citep{ellenberger2025backpropagation}. Recent work has also identified cellular mechanisms for prospective and retrospective firing~\citep{brandt2024prospective} and framed prospective neurons as a solution to teaching-signal synchronization~\citep{zucchet2025teaching}. Our work complements these circuit theories by showing that prospectivity can also serve as an input-side architectural addition for deep recurrent networks.

\noindent\textbf{Normalization, ORGaNICs, and quadrature dynamics.} The RQF construction also belongs to a line of recurrent cortical circuit models in which dynamics, prediction, and normalization are coupled. \citet{heeger2017theory} modeled neural activity as a compromise between feedforward drive, feedback drive, and prior drive, with prediction over time implemented by recursive quadrature pairs. ORGaNICs extend this idea to gated recurrent neural integrator circuits with dynamic normalization~\citep{heeger2019oscillatory}, a canonical neural computation~\citep{carandini2012normalization}. Most prior ORGaNICs work has focused on stability and normalization~\citep{rawat2024unconditional,morone2026stabilization,pal2025hierarchical}; RQFs were also combined with ORGaNICs to model temporal-frequency selectivity and direction selectivity of neurons in primary visual cortex~\citep{heeger2020recurrent}. The RQF is the corresponding single-channel complex primitive, whose complex state carries the quadrature pair.

\noindent\textbf{Credit assignment and trainability.} The trainability problem we study is related to, but distinct from, the classical vanishing-gradient problem in RNNs, which arises from products of recurrent Jacobians through time~\citep{pascanu2013difficulty}. Online alternatives such as RTRL avoid reverse-time unrolling but introduce large sensitivity states~\citep{williams1989learning}; recent work makes this more tractable by exploiting independent recurrent modules across layers~\citep{zucchet2023online}. Equilibrium propagation, Lagrangian extensions, and Hamiltonian echo methods instead recover gradient information from the response or time reversal of physical dynamical systems~\citep{scellier2017equilibrium,massar2025equilibrium,lopez2023self,pourcel2025learning}. Our learning mode of spatial-only backpropagation removes reverse-time temporal credit but does not solve the separate weight-transport problem across layers~\citep{lillicrap2016random}.

\noindent\textbf{State-space models and discretization upgrades.} SSMs have made diagonal recurrent substrates practical for long sequences. S4 introduced efficient structured state-space layers~\citep{gu2021efficiently}, while DSS and S4D showed that complex diagonal state matrices can retain much of this performance with simpler implementations~\citep{gupta2022diagonal,gu2022parameterization}. S5 and LRU refine this family through MIMO scans and carefully parameterized linear recurrences~\citep{smith2022simplified,orvieto2023resurrecting}. Mamba addresses a different limitation by making discretized parameters input-dependent~\citep{gu2023mamba}. We deliberately keep the RQF parameters input-independent so that the prospective-input intervention and its gradient scaling can be isolated; sRQFs relax this. Mamba-3 also uses two input taps, but its convex endpoint interpolation is analytically distinct from the signed prospective difference (Appendix~\ref{app:two_tap})~\citep{lahoti2026mamba}.

\section{Discussion}
\label{sec:discussion}

\noindent\textbf{Takeaways for machine learning.} Prospective-input coding is a two-tap change to a diagonal SSM's input path that adds no parameters and preserves the parallel scan. Prospective RQFs yield higher mean test accuracy than their matched controls in all configurations trained with full BPTT; the prospective $\alpha$P-S5 construction likewise exceeds native S5 in all comparisons. Separately, RQFs are a special case of diagonal SSMs with clear advantages: at width 32, a stack of these biologically motivated filters, carrying no normalization, no gating, and no residual path, is more accurate than S5 at every raw-audio depth while using fewer parameters.

\noindent\textbf{Takeaways for computational neuroscience.} Prospective firing has been measured in biological neurons, and the look-ahead operator that models it inverts a first-order membrane low-pass filter exactly. Our results give that inverse a specific role: carrying a teaching signal through a deep cascade of leaky integrators. The evidence is mechanistic rather than only behavioral. Under spatial-only backpropagation in a deep recurrent network, we measure the error gain surviving each backward hop across layers, and a prospective drive preserves several times more of it than an instantaneous one. The consequence is temporally local learning: a six-layer stack that never differentiates its recurrence through time achieves $83$\% on $16{,}000$-step raw audio, far longer than the sequences on which the local rules of NLA and GLE have been demonstrated.

\noindent\textbf{Limitations.} The prospective-input correction is not directly applicable to discrete-time gated RNNs such as LSTMs and GRUs~\citep{hochreiter1997long,chung2014empirical}. The operator $\chit = 1 + \tau\,\mathrm{d}/\mathrm{d}t$ is defined on a continuous-time trajectory and uses $\tau$ as an intrinsic integration timescale, whereas standard LSTMs and GRUs are specified by per-step algebraic update equations. A discrete architecture can be embedded in many possible continuous flows, but without such a substrate, there is no canonical derivative or effective time constant to invert. Our proof isolates the linear RQF substrate under spatial-only backpropagation. Nonlinear activations, residual connections, normalization layers, input-dependent parameters, and imperfect choices of the look-ahead scale can all change the size of the empirical gain. Finally, we did not attempt to maximize performance in this paper; future work should learn or adapt the effective time constants, use multi-timescale look-ahead operators, optimize the pole initialization, and test residual and normalization choices.

\noindent\textbf{Selective RQFs and future work.} The input gain alone raises Path-X accuracy by three to four points under both input rules (Table~\ref{tab:pathx_results}). Making $\gamma$ and $\omega$ input-dependent as well enables ceiling performance on Selective Copying 
(Table~\ref{tab:selective_copying}). Associative recall, state tracking, and language modeling would test whether prospective input complements selectivity beyond these tasks.

Spatial-only backpropagation isolates within-step spatial credit assignment but deliberately discards credit through recurrent time. Extending the prospective-input mechanism to local learning rules that retain useful temporal credit is therefore an open problem. Temporal credit assignment in RNNs is difficult because gradients propagated through time can vanish or explode~\citep{pascanu2013difficulty}; deriving local alternatives has consequently been a long-standing goal~\citep{williams1989learning,zucchet2023online,senn2024neuronal,ellenberger2025backpropagation}. Biological synapses do not receive gradients produced by unrolling an entire recurrent computation, yet the brain learns from temporally extended experience. The goal is to use neuroscience principles to identify recurrent dynamical models whose learning rules use only local information (local errors and activity) while retaining useful credit assignment through time.

\begin{ack}
We thank Stefano Martiniani, Jenny Listman, Kyle Stratis, Ruben Coen-Cagli, and Wayne Mackey for helpful discussions. This research was funded by Telepath. S.R.\ and M.M.\ are employees of Telepath. F.M.\ is a paid consultant for Telepath. D.J.H.\ is a cofounder of and paid consultant for Telepath. The authors declare no other competing interests.
\end{ack}

\bibliographystyle{unsrtnat}
\bibliography{references}

\newpage
\appendix
\numberwithin{equation}{section}

\section{Details on Recursive Quadrature Filters (RQFs)}
\label{app:rqf_details}

This appendix collects the technical material supporting the construction in Section~\ref{sec:rqf}. We first recover the leaky-integrator substrate of Eq.~\eqref{eq:leaky_integrator} by modeling $y$ as a dynamical process that minimizes a single-channel slice of the cortical energy function of~\citet{heeger2017theory} over time, and then collect the elementary linear dynamical systems facts that motivate the band-pass interpretation, the special-case relationship to SSMs, and the equivalence to a damped harmonic oscillator.

\subsection{Energy-based derivation of RQFs}
\label{app:energy_derivation}

The starting point is Eq.~1 of~\citet{heeger2017theory}, which includes a feedforward residual and a prior residual at every layer:
\begin{equation}
E_H=\sum_i\alpha_H^{(i)}\!\left[
\lambda_H^{(i)}\sum_j\bigl(y_j^{(i)}-z_j^{(i)}\bigr)^2
+\bigl(1-\lambda_H^{(i)}\bigr)\sum_j\bigl(y_j^{(i)}-\widehat y_j^{(i)}\bigr)^2
\right].
\label{eq:heeger_energy}
\end{equation}
We select one response coordinate, omit the terms contributed by the next layer, and condition on the supplied feedforward target $z$ and prior target $\widehat y$. We extend this scalar compromise to two real quadrature coordinates with shared parameters and encode them as $y=y_1+\iota y_2$ (and likewise for $z$ and $\widehat y$). The sum of the paired real squared residuals is then a squared complex modulus. Although Heeger's $\alpha_H(t)$ and $\lambda_H(t)$ may vary, the RQF construction holds them fixed during relaxation, takes $\alpha_H$ to be common to a quadrature pair, and identifies a learned per-channel $\gamma=\lambda_H$ (the subscript distinguishes Heeger's symbols from the RQF decay rate $\alpha$ and pole $\lambda$ used elsewhere). A channel-dependent $\alpha_H$ would instead induce channel-dependent effective base time constants.

Let $\tau_H$ denote Heeger's relaxation time. Gradient descent on the selected quadrature pair gives
\begin{equation}
\tau_H\dot y=-2\alpha_H\left[\gamma(y-z)+(1-\gamma)(y-\widehat y)\right].
\label{eq:heeger_pair_dynamics}
\end{equation}
Defining $\tau\triangleq\tau_H/(2\alpha_H)$ absorbs the common scale. Equivalently, the normalized pair energy is
\begin{equation}
E=\gamma|y-z|^2+(1-\gamma)|y-\widehat y|^2,
\qquad
\frac{\partial E}{\partial y^\dagger}=\gamma(y-z)+(1-\gamma)(y-\widehat y),
\label{eq:rqf_energy}
\end{equation}
and, with $z$ and $\widehat y$ held fixed, its Wirtinger-gradient dynamics $\tau\dot y=-\partial E/\partial y^\dagger$ are
\begin{equation}
\tau\dot y=-y+\gamma z+(1-\gamma)\widehat y.
\label{eq:rqf_leaky_recovered}
\end{equation}
Along this flow,
\[
\frac{\mathrm{d}E}{\mathrm{d}t}
=\frac{\partial E}{\partial y}\dot y
+\frac{\partial E}{\partial y^\dagger}\dot y^\dagger
=-\frac{2}{\tau}\left|\frac{\partial E}{\partial y^\dagger}\right|^2\leq 0.
\]
Thus, the energy is nonincreasing and decreases strictly away from stationary points, while the dynamics recover Eq.~\eqref{eq:leaky_integrator}.

The recurrent prior ($\widehat y$) comes from the quadrature predictor in Eq.~3 of~\citet{heeger2017theory}. For angular frequency $\Omega$, its real-coordinate form is
\begin{equation}
\begin{bmatrix}\widehat y_1(t)\\ \widehat y_2(t)\end{bmatrix}
=
\begin{bmatrix}
\cos(\Omega\Delta t)&-\sin(\Omega\Delta t)\\
\sin(\Omega\Delta t)& \cos(\Omega\Delta t)
\end{bmatrix}
\begin{bmatrix}y_1(t-\Delta t)\\y_2(t-\Delta t)\end{bmatrix},
\label{eq:heeger_rotation}
\end{equation}
or $\widehat y(t)=e^{\iota\Omega\Delta t}y(t-\Delta t)$: the state sampled $\Delta t$ earlier is rotated forward to predict the present. The RQF applies the same rotation to the present state to predict one look-ahead horizon ahead, $\widehat y=e^{\iota\omega\tau}y$ with $\omega=\Omega$ and $\Delta t=\tau$, and keeps the first-order term
\begin{equation}
e^{\iota\omega\tau}y
=\left[1+\iota\omega\tau+O\!\left((\omega\tau)^2\right)\right]y,
\qquad
\widehat y\triangleq\left.\chit(y)\right|_{\dot y=\iota\omega y}=(1+\iota\omega\tau)y.
\label{eq:heeger_first_order_closure}
\end{equation}
Substitution into Eq.~\eqref{eq:rqf_leaky_recovered} yields
\[
\tau\dot y=-y+\gamma z+(1-\gamma)(1+\iota\omega\tau)y,
\]
which recovers the RQF in Eq.~\eqref{eq:rqf_scalar}.

\subsection{State-space form, oscillator equivalence, and cascade bandwidth}
\label{app:linear_systems}

\noindent\textbf{Two-dimensional real state-space form.} We write the scalar complex-valued equation in Eq.~\eqref{eq:rqf_scalar} as a real two-dimensional coupled dynamical system to show that the RQF \textit{is} a damped harmonic oscillator.
Decomposing the complex state $y = u + \iota v$ and the input $z = z_R + \iota z_I$ into their real and imaginary parts reveals that Eq.~\eqref{eq:rqf_scalar} is equivalent to the two-dimensional real linear system
\begin{equation}
    \dot u \;=\; -\alpha u - \omega_0 v + \alpha z_R, \qquad
    \dot v \;=\; \omega_0 u - \alpha v + \alpha z_I,
    \label{eq:ssm_real}
\end{equation}
or, in matrix form, $\dot{\mathbf{x}} = \bA \mathbf{x} + \bB \bz$ with $\mathbf{x} = (u, v)^\top$, $\bz = (z_R, z_I)^\top$, $\bA = -\alpha \mathbf{I} + \omega_0 \mathbf{J}$, $\bB = \alpha \mathbf{I}$, and $\mathbf{J} = \big[\begin{smallmatrix} 0 & -1 \\ 1 & 0 \end{smallmatrix}\big]$, the standard planar rotation generator. The matrix $\bA$ therefore implements an isotropic exponential decay at rate $\alpha$ combined with a planar rotation at angular velocity $\omega_0$. The eigenvalues follow from $\det(\bA - \lambda \mathbf{I}) = (\lambda + \alpha)^2 + \omega_0^2 = 0$, giving the complex conjugate pair $\lambda_{1,2} = -\alpha \pm \iota\omega_0$; both lie strictly in the open left half-plane for any $\gamma \in (0,1)$, so the filter is unconditionally stable. The free response is $\mathbf{x}(t) = e^{-\alpha t}\,\mathbf{R}(\omega_0 t)\,\mathbf{x}(0)$ with $\mathbf{R}(\theta)$ a planar rotation, and the impulse response of the scalar complex form is the damped complex exponential $h(t) = \alpha\,e^{-\alpha t}(\cos(\omega_0 t) + \iota \sin(\omega_0 t))$ for $t \ge 0$, the same primitive that diagonal complex SSMs convolve with their input sequence.

\noindent\textbf{Frequency response and quality factor.} The band-pass interpretation follows directly from the scalar form $\dot y = \lambda y + \alpha z$. With a zero initial condition, the transfer function from $z$ to $y$ is $H(s) = \alpha/(s-\lambda)=\alpha/(s+\alpha-\iota\omega_0)$, so its magnitude on the imaginary axis is
\begin{equation}
    |H(\iota\Omega)| \;=\; \frac{\alpha}{\sqrt{\alpha^2 + (\Omega - \omega_0)^2}},
    \label{eq:lorentzian}
\end{equation}
whose square is a Lorentzian; it peaks at unity gain when $\Omega=\omega_0$ and has full half-power bandwidth $\Delta\Omega_{3\mathrm{dB}} = 2\alpha$. The two learnable parameters set the two axes of this response: $\omega$ sets the phase-prior frequency and, through $\omega_0=(1-\gamma)\omega$, the resonance location, while $\gamma$ sets the bandwidth through $\alpha=\gamma/\tau$. Combining them gives the band-pass quality factor
\begin{equation}
    Q \;=\; \frac{\omega_0}{2\alpha} \;=\; \frac{(1-\gamma)\,\omega\,\tau}{2\gamma}.
    \label{eq:Q}
\end{equation}
Small $\gamma$ therefore yields a narrow band-pass filter centered near the phase-prior frequency $\omega$, whereas large $\gamma$ broadens the gain toward first-order low-pass integration.

\noindent\textbf{Equivalence with a damped harmonic oscillator.} A scalar damped harmonic oscillator with natural (undamped) frequency $\omega_n$ and damping ratio $\zeta$ obeys the standard second-order equation
\begin{equation}
    \ddot x \;+\; 2 \zeta \omega_n\, \dot x \;+\; \omega_n^2\, x \;=\; F(t),
    \label{eq:dho_standard}
\end{equation}
where $F(t)$ is an external forcing; the damping ratio $\zeta$ controls the qualitative regime, with the system being undamped when $\zeta = 0$, underdamped when $0 < \zeta < 1$ (the impulse response is a decaying sinusoid), critically damped when $\zeta = 1$, and overdamped when $\zeta > 1$. To put the RQF in this form, we differentiate the first equation in Eq.~\eqref{eq:ssm_real}, substitute $\dot v$ from the second, and eliminate the residual $v$ using the first equation again. The result is a single second-order ODE in $u$,
\begin{equation}
    \ddot u \;+\; 2\alpha\, \dot u \;+\; (\alpha^2 + \omega_0^2)\, u \;=\; \alpha\bigl(\dot z_R + \alpha z_R\bigr) \;-\; \alpha \omega_0\, z_I,
    \label{eq:dho_from_rqf}
\end{equation}
which has exactly the form of Eq.~\eqref{eq:dho_standard} with the external forcing
\begin{equation}
    F(t) \;=\; \alpha\bigl(\dot z_R(t) + \alpha\, z_R(t)\bigr) \;-\; \alpha \omega_0\, z_I(t),
    \label{eq:dho_forcing}
\end{equation}
built from the in-phase and quadrature components of $z$: a proportional-plus-derivative response $\alpha(\dot z_R + \alpha z_R)$ to the real part, plus a static coupling $-\alpha\omega_0 z_I$ to the imaginary part. Factoring $\alpha$ out of the real-part term and introducing the effective time constant $\tau_0 \equiv 1/\alpha=\tau/\gamma$ exhibits the in-phase drive as a \emph{prospective} signal on the timescale $\tau_0$: $\alpha(\dot z_R + \alpha z_R) = \alpha^2 (z_R + \tau_0 \dot z_R) = \alpha^2\, \chi_{\tau_0}(z_R)$. The forcing can therefore be written compactly as
\begin{equation}
    F(t) \;=\; \alpha^2\, \chi_{\tau_0}(z_R(t)) \;-\; \alpha \omega_0\, z_I(t).
    \label{eq:dho_forcing_chi}
\end{equation}
Matching coefficients between Eqs.~\eqref{eq:dho_standard} and~\eqref{eq:dho_from_rqf} reads off the oscillator parameters as
\begin{equation}
    \omega_n \;=\; \sqrt{\alpha^2 + \omega_0^2} \;=\; |\lambda|, \qquad
    \zeta \;=\; \frac{\alpha}{\sqrt{\alpha^2 + \omega_0^2}} \;=\; \frac{\alpha}{|\lambda|},
    \label{eq:zeta_omega_n}
\end{equation}
and substituting the RQF parameterization $\alpha = \gamma/\tau$ and $\omega_0 = (1-\gamma)\omega$ gives the closed-form map from the learnable parameters $(\omega, \gamma, \tau)$ to the oscillator parameters,
\begin{equation}
    \omega_n \;=\; \frac{1}{\tau}\sqrt{\gamma^2 + (1-\gamma)^2\,\omega^2\tau^2}, \qquad
    \zeta \;=\; \frac{\gamma}{\sqrt{\gamma^2 + (1-\gamma)^2\,\omega^2\tau^2}}.
    \label{eq:zeta_omega_n_rqf}
\end{equation}
For any $\gamma \in (0,1)$ and $\omega\tau > 0$, the inequality $\zeta < 1$ holds strictly, so an RQF is always underdamped: its impulse response is a sinusoid at the damped oscillation frequency $\omega_0 = \omega_n\sqrt{1-\zeta^2}$ contained in an $e^{-\alpha t}$ envelope, and its squared magnitude response is the Lorentzian of Eq.~\eqref{eq:lorentzian}. The standard oscillator quality factor $Q_{\mathrm{osc}} = 1/(2\zeta) = \omega_n/(2\alpha)$ and the band-pass quality factor $Q = \omega_0/(2\alpha)$ from Eq.~\eqref{eq:Q} therefore differ by the factor $\omega_n/\omega_0 = 1/\sqrt{1-\zeta^2}$, which approaches unity in the high-$Q$ (low-$\zeta$) regime, and the two coincide as $\zeta \to 0$.

\noindent\textbf{Cascade bandwidth.} Here, we summarize the effect of cascading RQFs, which could be an architectural knob for adjusting per-channel selectivity.
A single RQF has a Lorentzian squared magnitude response with full half-power bandwidth $\mathrm{BW}_1 = 2\alpha$ set by Eq.~\eqref{eq:lorentzian}. Sharper per-channel selectivity at fixed $\alpha$ can be obtained by cascading $N$ identical RQFs \emph{within} a layer, i.e., in series with no nonlinearity between them. With $N$ identical RQFs in series, the transfer function is $H_N(s) = H(s)^N$, so by Eq.~\eqref{eq:lorentzian} the cascade magnitude squared is
\begin{equation}
    |H_N(\iota\Omega)|^2 \;=\; \frac{\alpha^{2N}}{\bigl(\alpha^2 + (\Omega - \omega_0)^2\bigr)^N},
    \label{eq:cascade_mag2}
\end{equation}
which peaks at unity at $\Omega = \omega_0$ for every $N$, since stacking unity-gain filters preserves unity gain at resonance. Setting $|H_N(\iota\Omega)|^2 = 1/2$ and solving for the frequency offset,
\begin{equation}
    \bigl(\alpha^2 + (\Omega - \omega_0)^2\bigr)^N \;=\; 2\,\alpha^{2N} \;\Longleftrightarrow\; (\Omega - \omega_0)^2 \;=\; \bigl(2^{1/N} - 1\bigr)\,\alpha^2,
    \label{eq:cascade_halfpower}
\end{equation}
gives the full half-power bandwidth $\mathrm{BW}_N = 2\alpha\sqrt{2^{1/N} - 1}$ and therefore the cascade bandwidth ratio
\begin{equation}
    \frac{\mathrm{BW}_N}{\mathrm{BW}_1} \;=\; \sqrt{2^{1/N} - 1} \;\xrightarrow[N \to \infty]{}\; \sqrt{\frac{\ln 2}{N}} \;\approx\; \frac{0.833}{\sqrt{N}},
    \label{eq:cascade_ratio}
\end{equation}
where the asymptotic form follows from $2^{1/N} = 1 + (\ln 2)/N + O(1/N^2)$. A four-fold within-layer cascade, for example, would narrow the per-channel bandwidth to $\sqrt{2^{1/4} - 1} \approx 0.435$ of a single-stage filter at fixed $\alpha$ while requiring $4\times$ the computation on the linear path and no extra learnable parameters (the cascade shares the channel's $(\omega_0, \alpha)$). We do not exploit this knob in the present work; every layer uses $N=1$, but it remains an inexpensive design choice.

\section{Discretization}
\label{app:discretization}

This appendix gives the full derivations of the ZOH discrete update in Eq.~\eqref{eq:zoh_main}, stated in Section~\ref{sec:rqf}, and the impulse-ZOH update for prospective-input coding that produces the two-tap input path referenced in Section~\ref{sec:method}. As in the main text, $\bz^{(\ell)}$ denotes the compact post-weight drive used for forward discretization; Appendix~\ref{app:gradient_proof} expands $\bz^{(\ell)}=\bW^{(\ell)}\bx^{(\ell)}$ when deriving learning rules.

\subsection{Instantaneous input}
\label{app:zoh}

Starting from the per-layer ODE $\dot{\by}^{(\ell)} = \blambda^{(\ell)} \odot \by^{(\ell)} + \balpha^{(\ell)} \odot \bz^{(\ell)}$ in Eq.~\eqref{eq:rqf_layer}, the element-wise decoupling reduces the vector ODE to $n_\ell$ independent scalar problems, each admitting the variation-of-constants solution over $[t_n, t_n + h]$:
\begin{equation}
    \by^{(\ell)}(t_n + h) \;=\; e^{\blambda^{(\ell)} h} \odot \by^{(\ell)}(t_n) \;+\; \balpha^{(\ell)} \odot \int_0^h e^{\blambda^{(\ell)}(h-s)} \odot \bz^{(\ell)}(t_n + s)\,\mathrm{d}s,
    \label{eq:voc_app}
\end{equation}
where the exponential is evaluated element-wise because $\blambda^{(\ell)}$ is a vector. The zero-order hold scheme replaces $\bz^{(\ell)}(t_n + s)$ on $(0, h]$ by the right-endpoint sample $\bz_{n+1}^{(\ell)} = \bW^{(\ell)}\rho(\by_{n+1}^{(\ell-1)})$, pulls it out of the integral, and evaluates the remaining kernel integral via $u = h - s$:
\begin{equation}
    \int_0^h e^{\blambda^{(\ell)}(h-s)}\,\mathrm{d}s \;=\; \int_0^h e^{\blambda^{(\ell)} u}\,\mathrm{d}u \;=\; \frac{e^{\blambda^{(\ell)} h} - \bone}{\blambda^{(\ell)}},
    \label{eq:kernel_integral_app}
\end{equation}
with division taken element-wise (well-defined because $\Real(\lambda_i^{(\ell)}) = -\alpha_i^{(\ell)} < 0$). Substituting the coefficients $\bar{\bA}_h^{(\ell)} \triangleq e^{\blambda^{(\ell)} h}$ and $\bar{\bB}_h^{(\ell)} \triangleq (\balpha^{(\ell)}/\blambda^{(\ell)}) \odot (\bar{\bA}_h^{(\ell)} - \bone)$ from Eq.~\eqref{eq:AB_main} yields the diagonal complex affine recurrence of Eq.~\eqref{eq:zoh_main}. Both coefficients depend only on the per-filter parameters $(\bgamma^{(\ell)}, \bomega^{(\ell)})$, the base time constant $\tau$, and the step size $h$, and can therefore be precomputed once per parameter update and reused across all time steps. The homogeneous recurrence is unconditionally stable, since $|e^{\lambda h}| = e^{-\alpha h} < 1$ for any $\alpha > 0$, and the ZOH update differs from forward Euler by $O((h/\tau)^2)$ for $h \ll \tau$.

\subsection{Prospective input}
\label{app:impulse_zoh}

For the prospective-input variant, applying $\chit$ to the layer input $\bx^{(\ell)}$ before $\bW^{(\ell)}$ is equivalent in the forward pass to driving Eq.~\eqref{eq:rqf_layer} by $\chit(\bz^{(\ell)}) = \bz^{(\ell)} + \tau\,\dot{\bz}^{(\ell)}$ in place of $\bz^{(\ell)}$. The variation-of-constants integral of Eq.~\eqref{eq:voc_app} therefore picks up an additional $\tau\,\dot{\bz}^{(\ell)}$ contribution. We model $\bz^{(\ell)}$ on $[t_n, t_n + h]$ as a right-continuous piecewise-constant signal that holds the value $\bz_{n+1}^{(\ell)}$ throughout the interval and that jumps from $\bz_n^{(\ell)}$ to $\bz_{n+1}^{(\ell)}$ at the left endpoint $s = 0$; this is the same hold convention used for the instantaneous-input ZOH update of Appendix~\ref{app:zoh}, made explicit at the boundary because the prospective term involves $\dot{\bz}^{(\ell)}$. Under this convention, $\dot{\bz}^{(\ell)}$ is a Dirac impulse at $s = 0$ of magnitude equal to the jump,
\begin{equation}
    \dot{\bz}^{(\ell)}(t_n + s) \;=\; \left(\bz_{n+1}^{(\ell)} - \bz_n^{(\ell)}\right) \delta(s),
    \label{eq:impulse_app}
\end{equation}
and the look-ahead $\chit(\bz^{(\ell)})$ splits additively into a constant part $\bz_{n+1}^{(\ell)}$ on $[0, h]$ and an impulsive part $\tau(\bz_{n+1}^{(\ell)} - \bz_n^{(\ell)})\delta(s)$ at $s = 0$. Each part contributes in closed form to the integral of Eq.~\eqref{eq:voc_app}. The constant part reproduces the calculation of Appendix~\ref{app:zoh} and contributes $\bar{\bB}_h^{(\ell)} \odot \bz_{n+1}^{(\ell)}$. The impulsive part is sampled by the kernel at $s = 0$, where $e^{\blambda^{(\ell)}(h-s)} = e^{\blambda^{(\ell)} h} = \bar{\bA}_h^{(\ell)}$ (we use the standard convention that the boundary impulse at $s = 0$ is included in $\int_0^h$). Multiplying the impulse strength by this kernel value and by the prefactor $\balpha^{(\ell)}$ from Eq.~\eqref{eq:voc_app} gives
\begin{equation}
    \balpha^{(\ell)}\,\tau \odot \bar{\bA}_h^{(\ell)} \odot \left(\bz_{n+1}^{(\ell)} - \bz_n^{(\ell)}\right) \;=\; \bgamma^{(\ell)} \odot \bar{\bA}_h^{(\ell)} \odot \left(\bz_{n+1}^{(\ell)} - \bz_n^{(\ell)}\right),
    \label{eq:impulse_contrib_app}
\end{equation}
where the equality uses $\balpha^{(\ell)}\tau = \bgamma^{(\ell)}$ from $\balpha^{(\ell)} = \bgamma^{(\ell)}/\tau$. Summing the two contributions and grouping by input sample yields the impulse-ZOH update of Eq.~\eqref{eq:zoh_prospective_main} in the main text. No finite-difference approximation has been introduced: the update is exact under the right-continuous piecewise-constant model of $\bz^{(\ell)}$.

\noindent\textbf{Computational cost.} Prospective input adds no trainable parameters to an RQF layer and leaves its diagonal state transition and associative scan unchanged. For a sequence of length $T$, layer $\ell$ maps $n_{\ell-1}$ inputs to $n_\ell$ RQF channels and computes the dense drive $\bz_t^{(\ell)}=\bW^{(\ell)}\bx_t^{(\ell)}$ in $O(Tn_\ell n_{\ell-1})$ work under either input rule. Prospective-input coding adds $O(Tn_\ell)$ element-wise computation and one lagged read of the already materialized drive, but no additional full-sequence activation storage; streaming execution needs only an $O(n_\ell)$ rolling buffer for the previous drive. The diagonal RQF scan remains $O(Tn_\ell)$ with $O(\log T)$ parallel depth. Thus, the asymptotic complexity of both inference and full BPTT is unchanged, although the second tap adds a small lower-order operation.

\subsection{Two-tap form for diagonal SSMs}
\label{app:two_tap}

The preceding derivation does not rely on the RQF parameterization. To transfer the prospective-input correction to the diagonal SSM core of Eq.~\eqref{eq:ssm_diagonal_core}, we keep the SSM parameters fixed and replace the input signal entering the continuous-time state equation by its look-ahead:
\begin{equation}
    \dot{\widetilde{\by}}(t)
    \;=\;
    \bLambda\,\widetilde{\by}(t)
    +
    \bB\,\chit(\bx(t))
    \;=\;
    \bLambda\,\widetilde{\by}(t)
    +
    \bB\!\left[\bx(t) + \tau\,\dot{\bx}(t)\right],
    \quad
    \bo(t) \;=\; \bC\,\widetilde{\by}(t) + \bD\,\bx(t),
    \label{eq:ssm_prospective_continuous}
\end{equation}
where $\tau$ is the look-ahead scale in $\chit$ and $(\bLambda,\bB,\bC,\bD)$ are exactly the SSM parameters from Eq.~\eqref{eq:ssm_diagonal_core}. Thus, the intervention is not a reparameterization of the SSM input matrix: $\bB$ remains the continuous-time input matrix, and the only change is that the state receives $\chit(\bx)$ instead of $\bx$. The feedthrough path is written with $\bx(t)$ because the prospective-input correction concerns the recurrent state input; applying the same look-ahead to the direct path would be a separate architectural choice.

Under the same right-continuous hold convention used in Appendix~\ref{app:impulse_zoh}, $\bx(t_n+s)$ takes the value $\bx_{n+1}$ on $[0,h)$ and has a jump from $\bx_n$ to $\bx_{n+1}$ at $s=0$. Therefore, $\dot{\bx}(t_n+s)=(\bx_{n+1}-\bx_n)\delta(s)$, and the variation-of-constants formula gives
\begin{equation}
    \widetilde{\by}_{n+1}
    \;=\;
    \bar{\bA}_h\,\widetilde{\by}_n
    +
    \bar{\bB}_h\,\bx_{n+1}
    +
    \tau\,\bar{\bA}_h\bB\,(\bx_{n+1}-\bx_n),
    \label{eq:ssm_prospective_impulse_zoh}
\end{equation}
where $\bar{\bA}_h=e^{\bLambda h}$ and $\bar{\bB}_h=\bLambda^{-1}(\bar{\bA}_h-\bI)\bB$ are the standard ZOH coefficients from Eq.~\eqref{eq:ssm_zoh_bar}. Grouping terms by input sample yields the two-tap input path
\begin{equation}
    \widetilde{\by}_{n+1}
    \;=\;
    \bar{\bA}_h\,\widetilde{\by}_n
    +
    \bar{\bB}_+\,\bx_{n+1}
    +
    \bar{\bB}_-\,\bx_n,
    \qquad
    \bar{\bB}_+
    \;\triangleq\;
    \bar{\bB}_h+\tau\,\bar{\bA}_h\bB,
    \qquad
    \bar{\bB}_-
    \;\triangleq\;
    -\tau\,\bar{\bA}_h\bB .
    \label{eq:ssm_two_tap}
\end{equation}
Eq.~\eqref{eq:ssm_two_tap} is the direct SSM analogue of Eq.~\eqref{eq:zoh_prospective_main}. The recurrent dynamics are unchanged: the state is advanced by the same diagonal transition $\bar{\bA}_h$, while the input enters through a two-tap feedforward path on the current and previous samples. For S4D, this amounts to replacing the one-tap ZOH input coefficient $\bar{\bB}_h$ by the pair $(\bar{\bB}_+,\bar{\bB}_-)$, whose values are determined by the original continuous-time input matrix $\bB$, the transition $\bar{\bA}_h$, and the look-ahead scale $\tau$; the pole parameterization, state transition, output projection, and feedthrough term are otherwise unchanged. The same coefficient replacement applies to any diagonal state-space model whose continuous-time core has the form of Eq.~\eqref{eq:ssm_diagonal_core}.

\noindent\textbf{Comparison with Mamba-3.} This two-tap form permits a concrete comparison with the exponential-trapezoidal rule of Mamba-3~\citep{lahoti2026mamba}. Writing $\mathbf{v}_t=\bB_t\bx_t$, that rule uses (in Mamba-3's notation)
\begin{equation}
    \mathbf{h}_t=\balpha_t\odot\mathbf{h}_{t-1}
    +(1-\lambda_t)\Delta_t\,\balpha_t\odot\mathbf{v}_{t-1}
    +\lambda_t\Delta_t\,\mathbf{v}_t,
    \qquad \lambda_t\in[0,1],
    \label{eq:mamba3_two_tap}
\end{equation}
so its endpoint weights form a convex blend. Prospective coding instead discretizes $\bB[\bx(t)+\tau\dot{\bx}(t)]$ and gives the signed previous-input tap $\bar{\bB}_-\bx_n=-\tau\bar{\bA}_h\bB\bx_n$ of Eq.~\eqref{eq:ssm_two_tap}. For a matched scalar real LTI system with $\Delta_t=h$, matching even this previous-input coefficient in Eq.~\eqref{eq:mamba3_two_tap} would require $\lambda_t=1+\tau/h>1$. The constructions therefore share a two-tap form but are not equivalent, and prospective input is not a special case of Mamba-3's admissible interpolation.

\section{Spatial-only backpropagation gradient derivation: full proof}
\label{app:gradient_proof}

This appendix gives the gradient calculation supporting Section~\ref{sec:method}. We consider an arbitrary real-valued loss local in time,
\begin{equation}
    \mathcal{L}_{k+1} \;=\; \Phi_{k+1}\!\left(\by^{(L)}_{k+1}, \by^{(L)\dagger}_{k+1}\right),
    \label{eq:local_time_loss_app}
\end{equation}
where $\Phi_{k+1}$ may include labels, readouts, or other quantities sampled at the same time step but does not backpropagate through earlier recurrent states. Spatial-only backpropagation freezes $\by^{(\ell)}_k$ for every layer $\ell$ and differentiates only through the within-step spatial chain $\by^{(0)}_{k+1}\to\cdots\to\by^{(L)}_{k+1}$. Unlike the forward-discretization derivation, we expand the compact drive as $\bz^{(\ell)}=\bW^{(\ell)}\bx^{(\ell)}$ because the gradients must pass through $\bW^{(\ell)}$. We write $\bx^{(\ell)}_{k+1}=\rho(\by^{(\ell-1)}_{k+1})$ and $\bx^{(\ell)}_k=\rho(\by^{(\ell-1)}_k)$, with $\bx^{(1)}$ supplied by the external input.

Because $\rho$ may be non-holomorphic, the exact chain rule is most transparent in real coordinates. For a complex vector and matrix, respectively, define
\begin{equation}
    \mathcal{R}(\mathbf{v})
    \;\triangleq\;
    \begin{bmatrix}
        \Real(\mathbf{v}) \\
        \Imag(\mathbf{v})
    \end{bmatrix},
    \qquad
    \mathcal{R}(\bW)
    \;\triangleq\;
    \begin{bmatrix}
        \Real(\bW) & -\Imag(\bW) \\
        \Imag(\bW) & \Real(\bW)
    \end{bmatrix},
    \label{eq:realification_app}
\end{equation}
so that $\mathcal{R}(\bW\mathbf{v})=\mathcal{R}(\bW)\mathcal{R}(\mathbf{v})$. For a real vector $\mathbf{g}$, let $\mathcal{D}(\mathbf{g})\triangleq\mathcal{R}(\operatorname{diag}(\mathbf{g}))$, and define the real activation Jacobian and layer error by
\begin{equation}
    \mathbf{J}_{\rho}(\by)
    \;\triangleq\;
    \frac{\partial\mathcal{R}(\rho(\by))}
         {\partial\mathcal{R}(\by)},
    \qquad
    \bdelta_{\R,k+1}^{(\ell)}
    \;\triangleq\;
    \nabla_{\mathcal{R}(\by_{k+1}^{(\ell)})}\mathcal{L}_{k+1}.
    \label{eq:delta_def_app}
\end{equation}
For split-ReLU, $\mathbf{J}_{\rho}$ is diagonal, with the real and imaginary activity masks on its two diagonal blocks, and $\|\mathbf{J}_{\rho}\|_2\leq 1$ (using any standard subgradient at zero). Moreover, $\|\mathcal{R}(\bW)\|_2=\|\bW\|_2$ and $\|\mathcal{D}(\mathbf{g})\|_2=\|\mathbf{g}\|_\infty$.
For clarity, we use forward Euler in the derivation because it exposes the factor of $h/\tau$ directly. The ZOH implementation has the same leading-order spatial gradient coefficients in the small-step limit: $\bar{\bA}_h^{(\ell)}=\bone+\blambda^{(\ell)}h+O(h^2)$ and $\bar{\bB}_h^{(\ell)}=(h/\tau)\bgamma^{(\ell)}+O(h^2)$, while the impulse-ZOH prospective coefficients satisfy $\bar{\bB}_h^{(\ell)}+\bgamma^{(\ell)}\odot\bar{\bA}_h^{(\ell)}=\bgamma^{(\ell)}+O(h)$ and $-\bgamma^{(\ell)}\odot\bar{\bA}_h^{(\ell)}=-\bgamma^{(\ell)}+O(h)$. Thus, the Euler calculation below proves the leading-order ZOH scaling as $h/\tau\to 0$.

\subsection{Spatial-only backward pass, instantaneous input}
\label{app:spatial_only_instantaneous}

Forward Euler discretization of the instantaneous-input RQF layer is
\begin{equation}
    \by^{(\ell)}_{k+1}
    \;=\;
    \bA^{(\ell)}\odot\by^{(\ell)}_k
    +
    \mathbf{g}_{\mathrm{np}}^{(\ell)}\odot\left(\bW^{(\ell)}\bx^{(\ell)}_{k+1}\right),
    \qquad
    \mathbf{g}_{\mathrm{np}}^{(\ell)}
    \;\triangleq\;
    \frac{h}{\tau}\,\bgamma^{(\ell)},
    \label{eq:np_forward_app}
\end{equation}
where $\bA^{(\ell)}=\bone+h\blambda^{(\ell)}$. Since $\by^{(\ell)}_k$ is frozen, the only live path from layer $\ell$ to layer $\ell+1$ runs through $\bx^{(\ell+1)}_{k+1}=\rho(\by^{(\ell)}_{k+1})$ and then through the matrix $\bW^{(\ell+1)}$. Applying the chain rule gives, for $\ell<L$,
\begin{equation}
    \bdelta_{\R,k+1}^{(\ell)}
    \;=\;
    \mathbf{J}_{\rho}\!\left(\by^{(\ell)}_{k+1}\right)^{\!\top}
    \mathcal{R}\!\left(\bW^{(\ell+1)}\right)^{\!\top}
    \mathcal{D}\!\left(\mathbf{g}_{\mathrm{np}}^{(\ell+1)}\right)^{\!\top}
    \bdelta_{\R,k+1}^{(\ell+1)} .
    \label{eq:np_delta_recursion_app}
\end{equation}
Consequently, in the Euclidean norm,
\begin{equation}
    \left\|\bdelta_{\R,k+1}^{(\ell)}\right\|_2
    \;\leq\;
    \prod_{m=\ell+1}^{L}
    \left(
        \left\|\mathbf{J}_{\rho}\!\left(\by^{(m-1)}_{k+1}\right)\right\|_2
        \left\|\bW^{(m)}\right\|_2
        \left\|\mathbf{g}_{\mathrm{np}}^{(m)}\right\|_\infty
    \right)
    \left\|\bdelta_{\R,k+1}^{(L)}\right\|_2 .
    \label{eq:np_delta_bound_app}
\end{equation}
When the weight norms, activation-Jacobian norms, and entries of $\bgamma^{(m)}$ are $O(1)$, Eq.~\eqref{eq:np_delta_bound_app} contains the explicit factor $(h/\tau)^{L-\ell}$. This is the discretization-induced depth attenuation: it is separate from any ordinary conditioning effects caused by the weights or nonlinearities. To state the parameter gradient without ambiguity about Wirtinger factors, let $\mathcal{C}$ invert vector realification, $\mathcal{C}([\mathbf{u}^{\top},\mathbf{v}^{\top}]^{\top})=\mathbf{u}+\iota\mathbf{v}$, and set $\bdelta_{\C,k+1}^{(\ell)}=\mathcal{C}(\bdelta_{\R,k+1}^{(\ell)})$. Define the packed complex gradient
\begin{equation}
    \nabla_{\bW}^{\mathrm{pack}}\mathcal{L}
    \;\triangleq\;
    \nabla_{\Real(\bW)}\mathcal{L}
    +
    \iota\,\nabla_{\Imag(\bW)}\mathcal{L}
    \;=\;
    2\,\frac{\partial\mathcal{L}}{\partial\bW^{*}} .
    \label{eq:packed_complex_gradient_app}
\end{equation}
The corresponding local gradient contribution has the outer-product form
\begin{equation}
    \nabla_{\bW^{(\ell)}}^{\mathrm{pack}}\mathcal{L}_{k+1}
    \;=\;
    \left(
        \mathbf{g}_{\mathrm{np}}^{(\ell)}
        \odot
        \bdelta_{\C,k+1}^{(\ell)}
    \right)
    \bx^{(\ell)\dagger}_{k+1}.
    \label{eq:np_weight_grad_app}
\end{equation}

\subsection{Spatial-only backward pass, prospective input}
\label{app:spatial_only_prospective}

For the prospective case, the look-ahead is applied to the layer input $\bx^{(\ell)}$ before multiplication by $\bW^{(\ell)}$. Using a backward difference at $t_{k+1}$ gives
\begin{equation}
    \widetilde{\bx}^{(\ell)}_{k+1}
    \;=\;
    \bx^{(\ell)}_{k+1}
    +
    \tau\,\frac{\bx^{(\ell)}_{k+1}-\bx^{(\ell)}_k}{h}
    \;=\;
    \frac{h+\tau}{h}\,\bx^{(\ell)}_{k+1}
    -
    \frac{\tau}{h}\,\bx^{(\ell)}_k .
    \label{eq:prospective_x_app}
\end{equation}
The forward Euler step is therefore
\begin{equation}
    \by^{(\ell)}_{k+1}
    \;=\;
    \bA^{(\ell)}\odot\by^{(\ell)}_k
    +
    \mathbf{g}_{+}^{(\ell)}
    \odot
    \left(\bW^{(\ell)}\bx^{(\ell)}_{k+1}\right)
    +
    \mathbf{g}_{-}^{(\ell)}
    \odot
    \left(\bW^{(\ell)}\bx^{(\ell)}_k\right),
    \label{eq:prospective_forward_app}
\end{equation}
with
\begin{equation}
    \mathbf{g}_{+}^{(\ell)}
    \;\triangleq\;
    \frac{h+\tau}{\tau}\,\bgamma^{(\ell)},
    \qquad
    \mathbf{g}_{-}^{(\ell)}
    \;\triangleq\;
    -\bgamma^{(\ell)} .
    \label{eq:prospective_gains_app}
\end{equation}
The previous input $\bx^{(\ell)}_k$ is frozen as a state variable for spatial error propagation, so it does not create a live path from $\by^{(\ell-1)}_{k+1}$ to $\by^{(\ell)}_{k+1}$. The current weight matrix multiplying that frozen input is still trainable. For the inter-layer error recursion, only the current-input coefficient $\mathbf{g}_{+}^{(\ell+1)}$ appears:
\begin{equation}
    \bdelta_{\R,k+1}^{(\ell)}
    \;=\;
    \mathbf{J}_{\rho}\!\left(\by^{(\ell)}_{k+1}\right)^{\!\top}
    \mathcal{R}\!\left(\bW^{(\ell+1)}\right)^{\!\top}
    \mathcal{D}\!\left(\mathbf{g}_{+}^{(\ell+1)}\right)^{\!\top}
    \bdelta_{\R,k+1}^{(\ell+1)} .
    \label{eq:prospective_delta_recursion_app}
\end{equation}
The corresponding norm bound is
\begin{equation}
    \left\|\bdelta_{\R,k+1}^{(\ell)}\right\|_2
    \;\leq\;
    \prod_{m=\ell+1}^{L}
    \left(
        \left\|\mathbf{J}_{\rho}\!\left(\by^{(m-1)}_{k+1}\right)\right\|_2
        \left\|\bW^{(m)}\right\|_2
        \left\|\mathbf{g}_{+}^{(m)}\right\|_\infty
    \right)
    \left\|\bdelta_{\R,k+1}^{(L)}\right\|_2 .
    \label{eq:prospective_delta_bound_app}
\end{equation}
Since $\|\mathbf{g}_{+}^{(m)}\|_\infty=(1+h/\tau)\|\bgamma^{(m)}\|_\infty$, the explicit small factor $(h/\tau)^{L-\ell}$ is absent. More precisely, at fixed depth and fixed parameters, the discretization prefactor on each hop is $O(1)$ rather than $O(h/\tau)$ as $h/\tau\to 0$. The bounds in Eqs.~\eqref{eq:np_delta_bound_app} and~\eqref{eq:prospective_delta_bound_app} are upper bounds and provide no lower bound: the remaining product of activation-Jacobian norms, weight norms, and $\bgamma$ can still attenuate with depth. We therefore make no claim that the complete prospective gradient is $O(1)$ or depth-independent. The local weight gradient contains both input taps:
\begin{equation}
    \nabla_{\bW^{(\ell)}}^{\mathrm{pack}}\mathcal{L}_{k+1}
    \;=\;
    \left(
        \mathbf{g}_{+}^{(\ell)}
        \odot
        \bdelta_{\C,k+1}^{(\ell)}
    \right)
    \bx^{(\ell)\dagger}_{k+1}
    +
    \left(
        \mathbf{g}_{-}^{(\ell)}
        \odot
        \bdelta_{\C,k+1}^{(\ell)}
    \right)
    \bx^{(\ell)\dagger}_{k}.
    \label{eq:prospective_weight_grad_app}
\end{equation}
Eq.~\eqref{eq:prospective_weight_grad_app} is local in time up to the one-step input memory introduced by the two-tap path, and local in space once the adjacent real-coordinate error $\bdelta_{\R,k+1}^{(\ell)}$ is available. Exact spatial backpropagation supplies that error through $\mathcal{R}(\bW^{(\ell+1)})^\top$ and the activation Jacobian in Eqs.~\eqref{eq:np_delta_recursion_app} and~\eqref{eq:prospective_delta_recursion_app}.

\section{Experimental results}
\label{app:results}

\subsection{Full BPTT results with 32 filters per layer}
\label{app:width32_results}

Tables~\ref{tab:rqf_bptt_width32} and~\ref{tab:s5_bptt_width32} report the width-32 results for RQF and $\alpha$P-S5, respectively.

\begin{table}[htbp]
    \centering
    \caption{\textbf{RQF, full BPTT.} Non-residual stacks with 32 filters per layer on Speech Commands.}
    \label{tab:rqf_bptt_width32}
    \setlength{\tabcolsep}{3.2pt}
    \begin{tabular}{c c cc c cc}
        \toprule
        & \multicolumn{3}{c}{Raw audio} & \multicolumn{3}{c}{MFCC} \\
        \cmidrule(lr){2-4}\cmidrule(lr){5-7}
        Depth & Params & Prospective & Non-pros. & Params & Prospective & Non-pros. \\
        \midrule
        2 & $23.5$k & $94.59\!\pm\!0.28$ & $93.96\!\pm\!0.28$ & $24.1$k & $96.38\!\pm\!0.13$ & $96.17\!\pm\!0.22$ \\
        4 & $27.7$k & $95.66\!\pm\!0.23$ & $95.29\!\pm\!0.22$ & $28.3$k & $96.33\!\pm\!0.23$ & $96.12\!\pm\!0.31$ \\
        6 & $31.9$k & $96.09\!\pm\!0.23$ & $95.27\!\pm\!0.22$ & $32.5$k & $96.35\!\pm\!0.17$ & $95.73\!\pm\!0.27$ \\
        \bottomrule
    \end{tabular}
\end{table}

\begin{table}[htbp]
    \centering
    \caption{\textbf{$\alpha$P-S5 versus native S5, full BPTT.} Residual S5 blocks with 32 units per layer on Speech Commands; the complete $\alpha$P construction changes the continuous-time input gain, adds the second tap, and uses the pole clipping described in Appendix~\ref{app:s5_architecture}.}
    \label{tab:s5_bptt_width32}
    \setlength{\tabcolsep}{3.2pt}
    \begin{tabular}{c c cc c cc}
        \toprule
        & \multicolumn{3}{c}{Raw audio} & \multicolumn{3}{c}{MFCC} \\
        \cmidrule(lr){2-4}\cmidrule(lr){5-7}
        Depth & Params & $\alpha$P-S5 & Native S5 & Params & $\alpha$P-S5 & Native S5 \\
        \midrule
        2 & $27.9$k & $93.46\!\pm\!0.93$ & $92.34\!\pm\!0.50$ & $28.6$k & $96.04\!\pm\!0.32$ & $95.36\!\pm\!0.24$ \\
        4 & $34.4$k & $94.74\!\pm\!0.59$ & $93.31\!\pm\!1.09$ & $35.1$k & $96.31\!\pm\!0.32$ & $95.84\!\pm\!0.32$ \\
        6 & $40.9$k & $95.06\!\pm\!0.99$ & $94.69\!\pm\!0.66$ & $41.5$k & $96.44\!\pm\!0.28$ & $95.84\!\pm\!0.18$ \\
        \bottomrule
    \end{tabular}
\end{table}

\subsection{Full BPTT results with residual connections}
\label{app:residual_results}

Table~\ref{tab:rqf_bptt_residual} repeats the full BPTT comparison of Table~\ref{tab:rqf_bptt}(a) with identity residual connections between layers. Each of the $L$ square RQF layers adds its input to its output through an identity connection, which adds no parameters; all other settings match Table~\ref{tab:rqf_bptt}(a). Every mean test accuracy is within $0.28$ points of its non-residual counterpart, and prospective input still gives the higher mean in every configuration.

\begin{table}[htbp]
    \centering
    \caption{\textbf{Residual RQF, full BPTT.} Stacks with 64 filters per layer and identity residual connections between layers on Speech Commands; otherwise identical to Table~\ref{tab:rqf_bptt}(a).}
    \label{tab:rqf_bptt_residual}
    \setlength{\tabcolsep}{3.2pt}
    \begin{tabular}{c c cc c cc}
        \toprule
        & \multicolumn{3}{c}{Raw audio} & \multicolumn{3}{c}{MFCC} \\
        \cmidrule(lr){2-4}\cmidrule(lr){5-7}
        Depth & Params & Prospective & Non-pros. & Params & Prospective & Non-pros. \\
        \midrule
        2 & $52.3$k & $94.84\!\pm\!0.25$ & $94.68\!\pm\!0.07$ & $53.5$k & $96.42\!\pm\!0.16$ & $96.30\!\pm\!0.27$ \\
        4 & $68.9$k & $96.08\!\pm\!0.18$ & $95.84\!\pm\!0.25$ & $70.2$k & $96.55\!\pm\!0.32$ & $96.49\!\pm\!0.23$ \\
        6 & $85.6$k & $96.27\!\pm\!0.24$ & $96.01\!\pm\!0.14$ & $86.8$k & $96.69\!\pm\!0.18$ & $96.54\!\pm\!0.11$ \\
        \bottomrule
    \end{tabular}
\end{table}

\subsection{Direct measurements of the spatial gradient path}
\label{app:gradient_measurements}

These measurements test how much of the error signal survives as it propagates backward through the layers. We use a six-layer, 64-channel raw-audio RQF under spatial-only backpropagation at the paper's operating point, $h/\tau=0.2$. Table~\ref{tab:gradient_norms} reports the weight-gradient norms: the two gradients are similar at layer 6, which is closest to the loss, but the instantaneous gradient decays much faster toward earlier layers. At layer 1, the prospective gradient is $10{,}290\times$ larger.

\begin{table}[htbp]
    \centering
    \caption{\textbf{Direct weight-gradient norms.} Frobenius norms $\|\nabla_{\bW^{(\ell)}}^{\mathrm{pack}}\mathcal{L}\|_F$ at $h/\tau=0.2$.}
    \label{tab:gradient_norms}
    \setlength{\tabcolsep}{9pt}
    \begin{tabular}{c c c c}
        \toprule
        Layer $\ell$ & Prospective & Instantaneous & Ratio \\
        \midrule
        1 & $7.69\times10^{-7}$ & $7.47\times10^{-11}$ & $10{,}290\times$ \\
        2 & $5.71\times10^{-6}$ & $6.01\times10^{-9}$  & $950\times$ \\
        3 & $1.03\times10^{-5}$ & $5.61\times10^{-8}$  & $185\times$ \\
        4 & $1.90\times10^{-5}$ & $5.62\times10^{-7}$  & $34\times$ \\
        5 & $3.08\times10^{-5}$ & $5.04\times10^{-6}$  & $6.1\times$ \\
        6 & $5.42\times10^{-5}$ & $4.87\times10^{-5}$  & $1.1\times$ \\
        \bottomrule
    \end{tabular}
\end{table}

Table~\ref{tab:gradient_sweep} sweeps the step size. We measure the per-hop error gain $g$ as follows: with $q_\ell$ the RMS norm of $\partial\mathcal{L}/\partial\bx^{(\ell)}$, we fit $g=(q_2/q_6)^{1/4}$ over the four hops from layer 6 to layer 2. The table compares the ratio $g_{\mathrm{pros}}/g_{\mathrm{inst}}$ with the exact-ZOH current-tap prediction $|\bar{\bB}_h+\bgamma\odot\bar{\bA}_h|/|\bar{\bB}_h|$ for the same filter bank. As $h/\tau$ decreases, $g_{\mathrm{pros}}$ remains between $0.585$ and $0.624$, whereas $g_{\mathrm{inst}}$ falls approximately in proportion to $h/\tau$. The measured ratio follows the predicted increase across the sweep.

\begin{table}[h!]
    \centering
    \caption{\textbf{Per-hop error gain versus step size.} $g_{\mathrm{pros}}$ and $g_{\mathrm{inst}}$ are the measured per-hop error gains for prospective and instantaneous input; prediction and measurement use the same six-layer, 64-channel filter bank.}
    \label{tab:gradient_sweep}
    \setlength{\tabcolsep}{7pt}
    \begin{tabular}{c c c c c}
        \toprule
        $h/\tau$ & Predicted ratio & $g_{\mathrm{pros}}$ & $g_{\mathrm{inst}}$ & Measured ratio \\
        \midrule
        $1.00$ & $1.96$  & $0.619$ & $0.361$  & $1.72$ \\
        $0.50$ & $2.95$  & $0.594$ & $0.220$  & $2.70$ \\
        $0.20$ & $5.93$  & $0.585$ & $0.104$  & $5.65$ \\
        $0.10$ & $10.91$ & $0.598$ & $0.0565$ & $10.59$ \\
        $0.05$ & $20.89$ & $0.624$ & $0.0305$ & $20.43$ \\
        \bottomrule
    \end{tabular}
\end{table}

\section{Architecture and hyperparameter details}
\label{app:hyperparams}

\subsection{RQF architecture}
\label{app:rqf_architecture}

A bias-free real projection maps the input dimension $d_{\mathrm{in}}\in\{1,20\}$ to width $n\in\{32,64\}$. This is followed by $L\in\{2,4,6\}$ square complex RQF layers of width $n$, with split-ReLU between layers and no normalization, gating, dropout, bidirectionality, or residual connections. At every step, the classifier concatenates the real and imaginary parts of the final state and applies a real MLP $2n\to256\to10$ with ReLU. Predictions use logits averaged across time. Parameter counts refer to real-valued trainable degrees of freedom, so a complex parameter counts twice. With $C=10$ classes, the count is
\begin{equation}
N_{\mathrm{params}}=d_{\mathrm{in}}n+2Ln^2+2Ln+512n+2826.
\label{eq:parameter_count}
\end{equation}

Raw waveforms have shape $(16{,}000,1)$ and use $h=1/16{,}000$ with fixed $\tau=5h=3.125\times10^{-4}$. MFCC inputs have shape $(161,20)$ and use their effective $160$~Hz frame clock, $h=1/160$ and $\tau=5h=0.03125$. Path-X uses $h=1/16{,}384$ and $\tau=5h$. No separate look-ahead scale is learned. Each RQF channel has complex state $s_i$, dimensionless frequency $\theta_i=\tau\omega_i$, bandwidth parameter $\gamma_i$, and continuous-time dynamics
\begin{equation}
    \frac{\mathrm{d}s_i}{\mathrm{d}t}
    \;=\;
    \lambda_i s_i
    +
    \frac{\gamma_i}{\tau} u_i(t),
    \qquad
    \lambda_i
    \;=\;
    -\frac{\gamma_i}{\tau}
    +
    \iota\,\omega_i(1-\gamma_i),
    \label{eq:rqf_channel_app}
\end{equation}
where $u_i(t)$ is the post-weight drive; $s_i$ and $u_i$ are the per-channel entries of $\by^{(\ell)}$ and $\bz^{(\ell)}$ in the main text. All runs use the ZOH recurrence in Eq.~\eqref{eq:rqf_zoh_app}; prospective input replaces its single input coefficient with the two-tap coefficients in Eq.~\eqref{eq:rqf_twotap_app}, with the delayed input initialized to zero.

The scalar instantaneous recurrence is
\begin{equation}
    s_{i,n+1}
    \;=\;
    \bar{A}_{h,i}s_{i,n}
    +
    \bar{B}_{h,i}u_{i,n+1},
    \qquad
    \bar{A}_{h,i}=e^{\lambda_i h},
    \qquad
    \bar{B}_{h,i}
    =
    \frac{\gamma_i}{\tau\lambda_i}
    \left(\bar{A}_{h,i}-1\right).
    \label{eq:rqf_zoh_app}
\end{equation}
The prospective-input recurrence keeps the same state coefficient and replaces the input contribution by the two-tap drive
\begin{equation}
    s_{i,n+1}
    \;=\;
    \bar{A}_{h,i}s_{i,n}
    +
    \left(\bar{B}_{h,i}+\gamma_i\bar{A}_{h,i}\right)u_{i,n+1}
    -
    \gamma_i\bar{A}_{h,i}u_{i,n},
    \label{eq:rqf_twotap_app}
\end{equation}
with the previous input buffer initialized to zero. This is the scalar-channel version of Eq.~\eqref{eq:zoh_prospective_main}.

\noindent\textbf{RQF parameterization and initialization.} The learned leaves are $\log\theta_i$ and $\log\gamma_i$:
\begin{equation}
    \theta_i=\tau\omega_i=e^{\log\theta_i},
    \qquad
    \gamma_i=e^{\log\gamma_i}.
    \label{eq:rqf_log_param}
\end{equation}
For a sequence with model-coordinate duration $T_{\mathrm{seq}}$, initialization is
\begin{equation}
    \log\theta_i\sim\mathcal{U}\!\left(\log\theta_{\min},\log\theta_{\max}\right),
    \quad
    \theta_{\min}=\frac{2\pi\tau}{T_{\mathrm{seq}}},
    \quad
    \theta_{\max}=2\pi F,
    \quad
    \gamma_i=\frac{\theta_i}{1+\theta_i}.
    \label{eq:rqf_initialization}
\end{equation}
All runs use $F=1$. Speech Commands uses $T_{\mathrm{seq}}=1$ in the model clock for each input representation, while Path-X uses $T_{\mathrm{seq}}=2\pi$. After every optimizer step, $\theta_i$ is clamped to $[\theta_{\min},2\pi]$ and $\gamma_i$ to $[10^{-6},1-10^{-6}]$.

\noindent\textbf{Weight and readout initialization.} Let $n_{\mathrm{in}}$ be the complex fan-in, $G_f=1/2$ the nominal RQF filter power gain, and $G_\rho$ the input activation's second-moment gain. Each real and imaginary component of a complex weight is sampled independently as follows:
\begin{equation}
\Real(W_{ij}),\Imag(W_{ij})\sim\mathcal{N}(0,\sigma_W^2),
\qquad
\sigma_W=\left(2n_{\mathrm{in}}G_fG_\rho\right)^{-1/2}.
\label{eq:rqf_weight_init}
\end{equation}
Thus, $G_\rho=1$ for a direct standardized input and $G_\rho=1/2$ after split-ReLU, giving $\sigma_W=1/\sqrt{n_{\mathrm{in}}}$ and $\sqrt{2/n_{\mathrm{in}}}$, respectively. The separate real input projection retains the default PyTorch linear initialization, uniform on $[-1/\sqrt{d_{\mathrm{in}}},1/\sqrt{d_{\mathrm{in}}}]$, and its bias is removed. The first classifier layer likewise retains its default initialization, with weights and biases uniform on $[-1/\sqrt{2n},1/\sqrt{2n}]$. The $256\to C$ logit weights are normal with standard deviation $10^{-3}$ and zero bias. The reference implementation evaluates the recurrence in complex64 with autocasting disabled; optimized scan kernels preserve the same coefficients.

\subsection{ORGaNICs architecture}
\label{app:organics_architecture}

The discrete ORGaNICs control follows the sequential model of~\citet{rawat2024unconditional}. We fix the recurrent matrix in every layer to $\mathbf{W}_r=\bI$, learn the intrinsic time constants, retain the gain and normalization dynamics, and stack $L\in\{2,4,6\}$ layers of width $64$. The reported parameter counts include the learned intrinsic time constants. Shape-matched layers use identity residual additions; the first $20\to64$ MFCC layer cannot use one. Layer normalization is disabled.

For layer $\ell$, let $\bx^{(\ell)}(t)\in\R^{d_\ell}$ denote its input, with $\bx^{(1)}$ equal to the 20-dimensional MFCC sequence. Each layer has 64-dimensional excitatory state $\by^{(\ell)}$, inhibitory normalization state $\mathbf{a}^{(\ell)}$, input gain $\mathbf{b}^{(\ell)}$, and semisaturation gain $\mathbf{b}_{0}^{(\ell)}$. We use the rectified-input variant, defining $[\mathbf{u}]_+=\max(\mathbf{u},\bzero)$ element-wise and
\begin{equation}
    \bz^{(\ell)}(t)
    \;=\;
    \left[\mathbf{W}_{zx}^{(\ell)}\bx^{(\ell)}(t)\right]_+,
    \qquad
    \bz_{\chi}^{(\ell)}(t)
    \;=\;
    \bz^{(\ell)}(t)
    +
    \delta_{\chi}\,
    \boldsymbol{\tau}_{y}^{(\ell)}
    \odot
    \dot{\bz}^{(\ell)}(t).
    \label{eq:organics_drive_app}
\end{equation}
Here, $\bz^{(\ell)}(t)$ is the input drive and $\bz_{\chi}^{(\ell)}(t)$ is the drive passed to the ORGaNICs input-gain pathway. The binary variable $\delta_{\chi}$ selects the input-drive condition: $\delta_{\chi}=0$ gives the instantaneous-input baseline, whereas $\delta_{\chi}=1$ gives the prospective-input variant $\chi_{\boldsymbol{\tau}_{y}^{(\ell)}}(\bz^{(\ell)}(t))$ with look-ahead scale $\boldsymbol{\tau}_{y}^{(\ell)}$.
The gain modulators evolve dynamically toward the following instantaneous targets:
\begin{equation}
    \begin{aligned}
    \mathbf{B}^{(\ell)}(t)
    &=
    \sigma_{\mathrm{sig}}\!\left(
        \mathbf{W}_{xb}^{(\ell)}\bx^{(\ell)}(t)
        +
        \mathbf{W}_{yb}^{(\ell)}\by^{(\ell)}(t)
        +
        \mathbf{W}_{ab}^{(\ell)}\mathbf{a}^{(\ell)}(t)
    \right), \\
    \mathbf{B}_{0}^{(\ell)}(t)
    &=
    \sigma_{\mathrm{sig}}\!\left(
        \mathbf{W}_{x0}^{(\ell)}\bx^{(\ell)}(t)
        +
        \mathbf{W}_{y0}^{(\ell)}\by^{(\ell)}(t)
        +
        \mathbf{W}_{a0}^{(\ell)}\mathbf{a}^{(\ell)}(t)
    \right),
    \end{aligned}
    \label{eq:organics_gain_app}
\end{equation}
where $\sigma_{\mathrm{sig}}$ is the logistic sigmoid; $\mathbf{B}^{(\ell)}$ is the target for the input gain $\mathbf{b}^{(\ell)}$; and $\mathbf{B}_{0}^{(\ell)}$ is the target for the semisaturation gain $\mathbf{b}_{0}^{(\ell)}$.

For the rectified-recurrence variant, define the recurrent prediction $\widehat{\by}^{(\ell)}=[\mathbf{W}_r\by^{(\ell)}]_+=[\by^{(\ell)}]_+$. The continuous-time ORGaNICs layer is then described by
\begin{equation}
\begin{aligned}
    \boldsymbol{\tau}_{y}^{(\ell)}
    \odot
    \dot{\by}^{(\ell)}
    &=
    -\by^{(\ell)}
    +
    \mathbf{b}^{(\ell)}
    \odot
    \bz_{\chi}^{(\ell)}
    +
    \left(
    \bone-
    \sqrt{\left[\mathbf{a}^{(\ell)}\right]_+}
    \right)
    \odot
    \widehat{\by}^{(\ell)}, \\
    \boldsymbol{\tau}_{a}^{(\ell)}
    \odot
    \dot{\mathbf{a}}^{(\ell)}
    &=
    -\mathbf{a}^{(\ell)}
    +
    \sigma_0^2
    \left(\mathbf{b}_{0}^{(\ell)}\right)^2
    +
    \mathbf{W}_{ay}^{(\ell)}
    \left(
        \left[\by^{(\ell)}\right]_+^2
        \odot
        \left[\mathbf{a}^{(\ell)}\right]_+
    \right), \\
    \boldsymbol{\tau}_{b}^{(\ell)}
    \odot
    \dot{\mathbf{b}}^{(\ell)}
    &=
    -\mathbf{b}^{(\ell)}
    +
    \mathbf{B}^{(\ell)}, \\
    \boldsymbol{\tau}_{b0}^{(\ell)}
    \odot
    \dot{\mathbf{b}}_{0}^{(\ell)}
    &=
    -\mathbf{b}_{0}^{(\ell)}
    +
    \mathbf{B}_{0}^{(\ell)} .
\end{aligned}
    \label{eq:organics_update_app}
\end{equation}
All products, powers, divisions by time constants, and square roots in Eq.~\eqref{eq:organics_update_app} are element-wise except for multiplication by learned matrices. In the implementation, Eq.~\eqref{eq:organics_update_app} is integrated with an ordered forward Euler step: the gain states are updated first, and those updated gains enter the principal and normalization updates. With $\dot{\bz}$ approximated by a backward difference, the Euler step multiplies the prospective term $\mathbf{b}\odot\boldsymbol{\tau}_{y}\odot(\bz_t-\bz_{t-1})/\Delta t$ by $\boldsymbol{\eta}_{y}=\Delta t/\boldsymbol{\tau}_{y}$, so the two-tap prospective correction reduces to $\mathbf{b}\odot(\bz_t-\bz_{t-1})$ in addition to the ordinary Euler drive, with the previous input-drive buffer initialized as $\bz_{-1}^{(\ell)}=\bzero$. The gain dynamics for $\mathbf{b}_0$ and $\mathbf{b}$ are otherwise identical in the prospective and instantaneous variants.

The normalization matrix $\mathbf{W}_{ay}^{(\ell)}$ is a learned full-rank positive matrix represented in log space; its initialization is 50\% sparse off-diagonal with unit diagonal. $\mathbf{W}_{zx}$ and the six gain-target matrices in Eq.~\eqref{eq:organics_gain_app} use Kaiming-uniform initialization with $a=\sqrt{5}$. The semisaturation parameter is fixed to $\sigma_0=1$. The four time-constant vectors are learned through bounded positive sigmoid maps; their unconstrained leaves are standard normal at initialization, with $0<\Delta t/\tau_y,\Delta t/\tau_a<0.05$ and $0<\Delta t/\tau_b,\Delta t/\tau_{b0}<0.1$. The principal state $\by^{(\ell)}(0)$ is a normalized random positive vector; $\mathbf{a}^{(\ell)}(0)\sim\mathcal{U}(0,1)$ is not normalized, $\mathbf{b}_{0}^{(\ell)}(0)=\bone$, and $\mathbf{b}^{(\ell)}(0)\sim\mathcal{U}(0,1)$. A learned linear $64\to10$ readout produces per-step logits, which are averaged for evaluation. The input correction is the only difference between the members of each prospective/non-prospective pair.

\subsection{\texorpdfstring{$\alpha$P-S5}{alphaP-S5} architecture and parameterization}
\label{app:s5_architecture}

The S5 controls use the native forward-only residual S5 architecture~\citep{smith2022simplified}. A real input projection maps $d_{\mathrm{in}}\in\{1,20\}$ to $n\in\{32,64\}$ units per layer. The $L\in\{2,4,6\}$ pre-normalized residual blocks use this value of $n$, conjugate symmetry, eight HiPPO blocks, ZOH discretization, and a half-GLU pointwise map. The readout normalizes the final $n$-dimensional block state and projects it to 64 dimensions; that representation is mean-pooled and classified by a $64\to256\to10$ GELU MLP with dropout $0.1$. The reported runs use batch pre-normalization and are unidirectional.

Let $\blambda$ collect the diagonal poles in S5's native clock, let $\bLambda=\mathrm{diag}(\blambda)$, and let $\boldsymbol{\Delta}$ collect the learned per-mode steps. For a physical input interval $h$, mode $p$ has effective continuous-time pole $\lambda_{\mathrm{eff},p}=(\Delta_p/h)\lambda_p$. We define $\alpha_p=-\Real(\lambda_p)>0$ and set $\bB_c=\mathrm{diag}(\balpha)\widetilde{\bB}$ in the native clock. The corresponding physical input row is $\bB_{\mathrm{eff},p:}=(\Delta_p/h)\bB_{c,p:}=-\Real(\lambda_{\mathrm{eff},p})\widetilde{\bB}_{p:}$, so the $\alpha$ gain remains tied to the physical pole decay rate and is not learned separately.

Prospective input uses the same shared physical horizon as the RQF experiments, $\tau=5h$. Substituting $\bB_{\mathrm{eff},p:}=(\Delta_p/h)\bB_{c,p:}$ into the impulse-ZOH correction of Eq.~\eqref{eq:ssm_prospective_impulse_zoh} gives $\tau\bar A_p\bB_{\mathrm{eff},p:}=5\Delta_p\bar A_p\bB_{c,p:}$. Thus, the factor $h$ cancels when the update is expressed in S5's native clock, and the implementation computes the following coefficients:
\begin{equation}
\begin{aligned}
\bar{\bA}
    &=\mathrm{diag}\!\left(e^{\blambda\odot\boldsymbol{\Delta}}\right), \\
\bar{\bB}
    &=\bLambda^{-1}(\bar{\bA}-\bI)\bB_c, \\
\bar{\bB}_{+}
    &=\bar{\bB}+5\,\mathrm{diag}(\boldsymbol{\Delta})\bar{\bA}\bB_c, \\
\bar{\bB}_{-}
    &=-5\,\mathrm{diag}(\boldsymbol{\Delta})\bar{\bA}\bB_c.
\end{aligned}
\label{eq:alpha_p_s5_taps}
\end{equation}
It then advances $\mathbf{h}_{t+1}=\bar{\bA}\mathbf{h}_t+\bar{\bB}_{+}\bx_{t+1}+\bar{\bB}_{-}\bx_t$. The two taps span adjacent input samples; $5\Delta_p$ is the native-clock impulse coefficient produced by the shared $5h$ horizon, not a learned or mode-dependent physical look-ahead. The delayed input starts at zero, and the direct $\bD\odot\bx_t$ feedthrough remains instantaneous. Within $\alpha$P-S5, the two-tap operation adds no trainable parameters and leaves the recurrent transition, residual path, output projection, and scan structure unchanged.

The poles and basis transformation use the standard HiPPO-DPLR initialization. Before transformation, entries of $\widetilde{\bB}$ and $\widetilde{\bC}$ are LeCun-normal with standard deviation $1/\sqrt{P_{\mathrm{local}}}$, and $D_i\sim\mathcal{N}(0,1)$. Log steps are initialized uniformly so that $\Delta_p\in[10^{-3},10^{-1}]$. In the reference $\alpha$P-S5 configuration, real pole parts are clipped to at most $-10^{-4}$ before forming $\balpha$, whereas the native one-tap S5 configuration does not apply this clipping and uses the unscaled learned input matrix. Tables~\ref{tab:s5_bptt} and~\ref{tab:s5_bptt_width32} therefore compare the complete $\alpha$P-S5 construction with native S5; they do not isolate the second tap while holding the continuous-time input map and pole clipping fixed.

\subsection{Optimizer and training}
\label{app:optimizer_training}

All Speech Commands runs use the 10-word subset of v0.02 with a stratified $70/15/15$ split (split seed $0$) and feature-wise standardization from the training split. MFCC extraction uses 20 coefficients, a 200-sample FFT window, 64 mel bands, and a 100-sample hop, producing 161 frames. A checkpoint is saved each epoch, and test accuracy is read from the checkpoint with the highest validation accuracy. Prospective/non-prospective pairs use identical training settings and seeds. Neither Speech Commands nor Path-X uses data augmentation. The Speech Commands v0.02 archive includes a CC BY~4.0 license. Path-X was obtained from the official LRA release; the accompanying code repository is Apache~2.0, but the downloaded data archive does not include separate dataset-license terms.

\paragraph{RQF.}
RQF Speech Commands models use AdamW with a batch size of $32$; the reported runs use seeds $42$--$46$. Complex mixing weights, the input projection, and the readout use a learning rate of $10^{-3}$ and weight decay $10^{-4}$; RQF pole parameters use a learning rate of $10^{-4}$ and zero weight decay. The global gradient norm is clipped at $1.0$. Training runs for at most 300 epochs with cosine annealing to $10^{-6}$ and stops after 20 epochs without improved validation accuracy. Full BPTT retains the complete temporal graph and applies cross entropy with label smoothing $0.1$ to the mean-pooled logits. Spatial-only backpropagation detaches the recurrent state after every step and applies the same smoothed cross entropy per time step; validation and test predictions average logits across time.

\paragraph{ORGaNICs.}
ORGaNICs models use AdamW with a learning rate of $10^{-3}$ and weight decay $10^{-4}$, cross entropy with label smoothing $0.1$, and gradient clipping at $1.0$. The learning rate follows cosine annealing to $10^{-6}$ over a maximum of 300 epochs, with early stopping after 20 epochs without improved validation accuracy. Full BPTT runs use the mean-pooled prediction loss; spatial-only backpropagation detaches every recurrent state after each step and applies the loss at each time step.

\paragraph{$\alpha$P-S5.}
S5 uses the same AdamW optimizer, base batch size, label smoothing, clipping, cosine schedule, early stopping, and checkpoint selection as the RQF Speech Commands runs; the reported runs use seeds $0$--$4$. All S5 parameters, including $\bLambda$, $\widetilde{\bB}$, $\widetilde{\bC}$, and the log steps, use the base $10^{-3}$ learning rate and $10^{-4}$ weight decay. The native-clock gain $\balpha=-\Real(\operatorname{diag}(\bLambda))$ introduces no optimizer state because it is derived from the poles.

\paragraph{Path-X.}
We use the $200{,}000$-example \texttt{pathfinder128/curv\_contour\_length\_14} task with a seeded $80/10/10$ split (split seed $42$). Pixels are divided by 255 and standardized by the training-split scalar mean and standard deviation. The six-layer, width-64 RQFs and sRQFs use Adam with $\epsilon=10^{-4}$, zero weight decay, a learning rate of $3\times10^{-3}$ for all parameter groups, a batch size of $64$, cross entropy at each time step, and gradient clipping at $1.0$. Training runs for 300 epochs with cosine annealing to zero; evaluation averages logits over time and reports the test result at the best-validation epoch. The five runs use seeds $0$--$4$ for initialization and data-shuffling order while retaining the same data split. Path-X uses the same recurrence, log-space parameterization, split-ReLU, and real--imaginary MLP readout as the Speech Commands models. The optimizer schedule and the initialization duration $T_{\mathrm{seq}}$ were chosen for Path-X rather than carried over from the Speech Commands runs. The sRQFs differ from the RQFs only by the input gain of Appendix~\ref{app:selective_rqf}; their $\gamma$ and $\omega$ are not input-dependent.

\section{Selective RQFs (sRQFs)}
\label{app:selective_rqf}

An sRQF layer (Eq.~\eqref{eq:srqf}) has the same drive $\bz^{(\ell)}_n=\bW^{(\ell)}\bx^{(\ell)}_n$ as an RQF layer and adds one input-dependent gain per filter, sampled at each step as
\begin{equation}
    \bm{\kappa}^{(\ell)}_n
    \;=\;
    \phi\!\left(\bW_{\kappa}^{(\ell)}
    \begin{bmatrix}\Real(\by^{(\ell-1)}_n)\\ \Imag(\by^{(\ell-1)}_n)\end{bmatrix}\right),
    \label{eq:srqf_gain}
\end{equation}
where $\bW_{\kappa}^{(\ell)}\in\R^{n_\ell\times 2n_{\ell-1}}$ is a learned, bias-free matrix and $\phi$ is a positive function applied element-wise. Our experiments use the example $\phi(u)=2\,\operatorname{sigmoid}(u)$, which lies in $(0,2)$ and equals $1$ at $u=0$. The gain reads the preceding layer's output before the activation $\rho$, so it retains the sign information that the rectified layer input $\bx^{(\ell)}_n=\rho(\by^{(\ell-1)}_n)$ discards; at the first layer, it reads the network input. Because the gain matrix is initialized at zero and this $\phi$ satisfies $\phi(0)=1$, $\bm{\kappa}^{(\ell)}_n=\bone$ at initialization and each sRQF starts as the corresponding RQF.

When $\bgamma$ and $\bomega$ are also selective, they are computed from the same features at each step:
\begin{equation}
\begin{aligned}
    \bgamma^{(\ell)}_n &= \gamma_{\min}+(\gamma_{\max}-\gamma_{\min})\,\operatorname{sigmoid}\!\left(\bm{\beta}^{(\ell)}+\bW_{\gamma}^{(\ell)}\begin{bmatrix}\Real(\by^{(\ell-1)}_n)\\ \Imag(\by^{(\ell-1)}_n)\end{bmatrix}\right),\\
    \bomega^{(\ell)}_n &= \bomega^{(\ell)}\odot\exp\!\left(c\,\tanh\!\left(\bW_{\omega}^{(\ell)}\begin{bmatrix}\Real(\by^{(\ell-1)}_n)\\ \Imag(\by^{(\ell-1)}_n)\end{bmatrix}\right)\right),
\end{aligned}
\label{eq:selective_rqf}
\end{equation}
where $\bW_{\gamma}^{(\ell)},\bW_{\omega}^{(\ell)}\in\R^{n_\ell\times 2n_{\ell-1}}$ are initialized at zero and $\bm{\beta}^{(\ell)}$ is the element-wise logit of $(\bgamma^{(\ell)}-\gamma_{\min})/(\gamma_{\max}-\gamma_{\min})$, so that each sRQF starts as the corresponding RQF; $c>0$ bounds each frequency multiplier to $[e^{-c},e^{c}]$. The Selective Copying sRQFs use $\gamma_{\min}=10^{-6}$, $\gamma_{\max}=1-10^{-6}$, and $c=\ln 4$, and split the filters of each layer by initial frequency into four equal groups that share one frequency multiplier, so $\bW_{\omega}^{(\ell)}$ has four distinct rows.

The selective quantities are held at $\bm{\kappa}^{(\ell)}_{n+1}$, $\bgamma^{(\ell)}_{n+1}$, and $\bomega^{(\ell)}_{n+1}$ over each interval $[t_n,t_{n+1}]$, the same right-endpoint convention as the ZOH drive. The gain then scales the input matrix of Eq.~\eqref{eq:rqf_as_ssm} to $\mathrm{diag}(\bm{\kappa}^{(\ell)}_{n+1})\,\mathrm{diag}(\balpha^{(\ell)}_{n+1})\bW^{(\ell)}$, and each interval is a fixed-parameter problem, so the derivations of Appendices~\ref{app:zoh} and~\ref{app:impulse_zoh} apply with the pole $\blambda_{n+1}=-\balpha_{n+1}+\iota(\bone-\bgamma_{n+1})\odot\bomega_{n+1}$ and $\balpha_{n+1}=\bgamma_{n+1}/\tau$. Suppressing the layer index, the instantaneous and prospective ZOH updates of Eqs.~\eqref{eq:zoh_main} and~\eqref{eq:zoh_prospective_main} become
\begin{align}
    \by_{n+1} &= \bar{\bA}_{n+1}\odot\by_n+\bm{\kappa}_{n+1}\odot\bar{\bB}_{n+1}\odot\bz_{n+1},
    \label{eq:srqf_instantaneous}\\
    \by_{n+1} &= \bar{\bA}_{n+1}\odot\by_n+\bm{\kappa}_{n+1}\odot\left[\left(\bar{\bB}_{n+1}+\bgamma_{n+1}\odot\bar{\bA}_{n+1}\right)\odot\bz_{n+1}-\bgamma_{n+1}\odot\bar{\bA}_{n+1}\odot\bz_n\right],
    \label{eq:srqf_prospective}
\end{align}
respectively, where $\bar{\bA}_{n+1}=e^{\blambda_{n+1}h}$ and $\bar{\bB}_{n+1}=(\balpha_{n+1}/\blambda_{n+1})\odot(\bar{\bA}_{n+1}-\bone)$. With input-independent $\bgamma$ and $\bomega$, as on Path-X, $\bar{\bA}_{n+1}=\bar{\bA}_h$ and $\bar{\bB}_{n+1}=\bar{\bB}_h$; setting $\bm{\kappa}_n=\bone$ as well recovers Eqs.~\eqref{eq:zoh_main} and~\eqref{eq:zoh_prospective_main}. The current gain multiplies both prospective taps: the look-ahead acts on the drive, and the gain scales the resulting input contribution, so the lagged drive $\bz_n$ is not multiplied by $\bm{\kappa}_n$; likewise, both taps use the current $\bgamma_{n+1}$ and $\bar{\bA}_{n+1}$. The gain alone changes how strongly each input is written into the state but not the pole, so a gain-only sRQF keeps the tuning frequency and bandwidth analyzed in Appendix~\ref{app:linear_systems}; selective $\bgamma$ and $\bomega$ instead move the pole at every step.

\noindent\textbf{Parallel scan.} The selective quantities depend only on the preceding layer's output and not on the layer's own state, so once that output is available, the transitions $\mathbf{a}_{n+1}=\bar{\bA}_{n+1}$ and the input terms $\mathbf{d}_{n+1}$ of Eqs.~\eqref{eq:srqf_instantaneous} and~\eqref{eq:srqf_prospective} can be computed for all samples in parallel. Both updates then take the form $\by_{n+1}=\mathbf{a}_{n+1}\odot\by_n+\mathbf{d}_{n+1}$, a diagonal linear recurrence whose coefficients may vary with $n$. Pairs $(\mathbf{a},\mathbf{d})$ compose associatively as $(\mathbf{a}_2,\mathbf{d}_2)\circ(\mathbf{a}_1,\mathbf{d}_1)=(\mathbf{a}_2\odot\mathbf{a}_1,\,\mathbf{d}_2+\mathbf{a}_2\odot\mathbf{d}_1)$ whether or not $\mathbf{a}$ varies with $n$, so the parallel scan for diagonal linear recurrences applies, with $O(Tn_\ell)$ work and $O(\log T)$ parallel depth~\citep{blelloch1990prefix,martin2018parallelizing,smith2022simplified}. This is the same property that lets Mamba compute its time-varying selective SSMs with a parallel scan~\citep{gu2023mamba}. As in Mamba, however, the time-varying transition removes the equivalence to a convolution~\citep{gu2023mamba}: FFT convolution remains available only for gain-only sRQFs, whose fixed $\bar{\bA}_h$ makes $\by$ a fixed-kernel causal convolution of $\mathbf{d}$. Each selective projection adds at most $O(Tn_\ell n_{\ell-1})$ work, the same order as the dense drive; $\bW_{\kappa}^{(\ell)}$ and $\bW_{\gamma}^{(\ell)}$ each add $2n_\ell n_{\ell-1}$ real parameters per layer, and the grouped $\bW_{\omega}^{(\ell)}$ adds $8n_{\ell-1}$. All sRQF results use full BPTT, which differentiates through the selective quantities; the spatial-only analysis of Appendix~\ref{app:gradient_proof} does not cover them.

\subsection{Selective Copying protocol}
\label{app:selective_copying_protocol}

The sRQFs make $\bgamma$, $\bomega$, and the input gain input-dependent (Appendix~\ref{app:selective_rqf}). Each sequence has $4{,}096$ one-hot inputs over a vocabulary of 16 symbols: blank token 0, data tokens 1--14, and recall marker 15. Sixteen distinct positions among the first $4{,}080$ are drawn uniformly and sorted, and each receives an independent uniformly drawn data token, so token values can repeat; all other context positions are blank. The final 16 inputs are recall markers, and their targets are the data tokens in input order. This follows the sequence length, vocabulary size, and number of data tokens of \citet{gu2023mamba}. Each epoch draws $64{,}000$ fresh training sequences; validation and test use separate fixed sets of $1{,}024$ sequences. The loss is the cross entropy averaged over the 16 recall positions, and the reported metric is recall-token accuracy, the fraction of recall tokens predicted correctly.

The RQFs and sRQFs both have four layers of 32 complex filters and no input projection: the first complex matrix maps the one-hot input directly to the filter bank. Between layers, the activation applies split-ReLU followed by divisive normalization across the filters of the layer:
\[
    [\rho(\by)]_i=\frac{\rho_0(y_i)}{\sqrt{\sigma^2+\frac{1}{n}\sum_{j=1}^{n}|\rho_0(y_j)|^2}},
\]
where $\rho_0$ is the split-ReLU of Eq.~\eqref{eq:split_relu}; $n=32$ is the layer width; and $\sigma=0.3$. There are no residual connections or dropout. At each recall position, the readout concatenates the real and imaginary parts of the last layer's state and applies a $64\to128\to16$ MLP with a ReLU hidden layer. The RQF parameterization and initialization follow Eqs.~\eqref{eq:rqf_log_param}--\eqref{eq:rqf_weight_init} with $h=1$, $\tau=5h$, $T_{\mathrm{seq}}=4{,}096h$, $F=1$, and $G_\rho=1$ both for the one-hot input and after divisive normalization. The readout's hidden layer is He-normal with zero bias, and its logit layer is normal with standard deviation $10^{-3}$ and zero bias. With complex parameters counted twice, the RQF and sRQF have $17{,}808$ and $33{,}040$ parameters, respectively.

All models are trained with full BPTT using Adam with a learning rate of $10^{-3}$ for all parameters, $\epsilon=10^{-4}$, and no weight decay; a batch size of 64; global gradient-norm clipping at $1.0$; and no label smoothing, learning-rate schedule, or early stopping. Training runs for 600 epochs of 1,000 updates. Validation accuracy is computed after every epoch, and test accuracy is read from the checkpoint with the highest validation accuracy. Each configuration is trained with five seeds, and the seed sets both the initialization and the data generation. Matched runs differ only in the input rule and in whether the gain, $\bgamma$, and $\bomega$ are input-dependent.

\end{document}